\documentclass{article}

\usepackage[final]{neurips_data_2024}

\pdfoutput=1
\usepackage{natbib}
\setcitestyle{numbers,square}
\usepackage[utf8]{inputenc} % allow utf-8 input
\usepackage[T1]{fontenc}    % use 8-bit T1 fonts
\usepackage{hyperref}       % hyperlinks
\usepackage{url}            % simple URL typesetting
\usepackage{booktabs}       % professional-quality tables
\usepackage{amsfonts}       % blackboard math symbols
\usepackage{nicefrac}       % compact symbols for 1/2, etc.
\usepackage{microtype}      % microtypography
\usepackage{xcolor}         % colors

\usepackage{subfigure}
\usepackage{paralist}
\usepackage{amsmath}
\usepackage{amssymb}
\usepackage{mathtools}
\usepackage{amsthm}
\usepackage{graphicx}
\usepackage{wrapfig, lipsum, booktabs}
\usepackage{color, colortbl}
\usepackage{bbm}
\usepackage{subcaption}
\usepackage{multirow}
\usepackage{tikz}
\usepackage{enumitem}
\usepackage[ruled,vlined]{algorithm2e}

\definecolor{citecol}{HTML}{2DDC0E}
\definecolor{tableofcontent}{HTML}{E63E15}
\definecolor{urlcol}{HTML}{2470D8}
\usepackage{hyperref}
\usepackage{multirow}
\hypersetup{
    colorlinks=true,       % false: boxed links; true: colored links
    linkcolor=tableofcontent,
    citecolor=citecol,        % color of links to bibliography
    urlcolor=black,           % color of external links
}
\usepackage{pifont}% http://ctan.org/pkg/pifont
\newcommand{\cmark}{\ding{51}}%
\newcommand{\xmark}{\ding{55}}%

\definecolor{Gray}{gray}{0.9}
\definecolor{Gray}{gray}{0.9}
\newcommand{\xhdr}[1]{{\vspace{1pt}\noindent\bfseries #1}.}
\newcommand{\ie}{\textit{i.e., }}
\newcommand{\eg}{\textit{e.g., }}

\begin{document}

\title{DTGB: A Comprehensive Benchmark for Dynamic Text-Attributed Graphs}

% The \author macro works with any number of authors. There are two commands
% used to separate the names and addresses of multiple authors: \And and \AND.
%
% Using \And between authors leaves it to LaTeX to determine where to break the
% lines. Using \AND forces a line break at that point. So, if LaTeX puts 3 of 4
% authors names on the first line, and the last on the second line, try using
% \AND instead of \And before the third author name.

\author{
Jiasheng Zhang$^1$ \quad Jialin Chen$^2$ \quad Menglin Yang$^2$
\quad Aosong Feng$^2$ \quad Shuang Liang$^1$\\ \textbf{Jie Shao}$^{1,3}$\thanks{Corresponding author.} \quad \textbf{Rex Ying}$^2$\\
$^1$University of Electronic Science and Technology of China \quad $^2$Yale University\\
$^3$Shenzhen Institute for Advanced Study, University of Electronic Science and Technology of China\\
\texttt{zjss12358@std.uestc.edu.cn} \quad \texttt{\{shuangliang, shaojie\}@uestc.edu.cn}\\
\texttt{\{jialin.chen, menglin.yang, aosong.feng,
rex.ying\}@yale.edu} }

\maketitle

\setcounter{footnote}{0}
\begin{abstract}
Dynamic text-attributed graphs (DyTAGs) are prevalent in various
real-world scenarios, where each node and edge are associated with
text descriptions, and both the graph structure and text
descriptions evolve over time. Despite their broad applicability,
there is a notable scarcity of benchmark datasets tailored to
DyTAGs, which hinders the potential advancement in many research
fields. To address this gap, we introduce \textbf{D}ynamic
\textbf{T}ext-attributed \textbf{G}raph \textbf{B}enchmark
(\textbf{DTGB}), a collection of large-scale, time-evolving graphs
from diverse domains, with nodes and edges enriched by dynamically
changing text attributes and categories. To facilitate the use of
DTGB, we design standardized evaluation procedures based on four
real-world use cases: future link prediction, destination node
retrieval, edge classification, and textual relation generation.
These tasks require models to understand both dynamic graph
structures and natural language, highlighting the unique challenges
posed by DyTAGs. Moreover, we conduct extensive benchmark
experiments on DTGB, evaluating 7 popular dynamic graph learning
algorithms and their variants of adapting to text attributes with
LLM embeddings, along with 6 powerful large language models (LLMs).
Our results show the limitations of existing models in handling
DyTAGs. Our analysis also demonstrates the utility of DTGB in
investigating the incorporation of structural and textual dynamics.
The proposed DTGB fosters research on DyTAGs and their broad
applications. It offers a comprehensive benchmark for evaluating and
advancing models to handle the interplay between dynamic graph
structures and natural language. The dataset and source code are
available at \url{https://github.com/zjs123/DTGB}.
\end{abstract}

\section{Introduction}
\label{sec:introduction}

Dynamic graphs are an essential tool for modeling a wide range of
real-world systems, such as e-commerce platforms, social networks,
and knowledge graphs \cite{tang2023dynamic, luo2023hope,
huang2022ttergm, zhang2023streame, cai2023temporal,
kazemi2020representation, skarding2021foundations}. In those dynamic
graphs, nodes and edges are typically associated with text
attributes, giving rise to dynamic text-attributed graphs (DyTAGs).
For example, e-commerce graphs may contain items accompanied by
textual descriptions, and time-annotated edges representing user
reviews or interactions with items. Similarly, temporal knowledge
graphs represent the sequential interactions among real-world
entities through textural relations. The exploration of learning
methodologies applied to DyTAGs is important to research areas such
as dynamic graph modeling and natural language processing
\cite{DyGLib, DGB, ye2024natural}, as well as various real-world
applications, \eg recommendation and social analysis
\cite{huang2022ttergm, khrabrov2010discovering, deng2019learning,
song2019session, zhang2022dynamic}.

While previous works for dynamic graph learning have proposed many
datasets describing the temporal interactions across different
domains \cite{kumar2019predicting, TGB, DGB, nath2023tboost}, these
datasets often lack edge categories and only contain statistical
features derived from raw attributes, lacking the raw text
descriptions of nodes and edges. Therefore, they fall short in
facilitating methodological advances in semantic modeling within
dynamic graphs and exploring the impact of text attributes on
downstream tasks. Concurrently, text-attributed graphs (TAGs) are
widely used in many real-world scenarios \cite{Microsoft,
yasunaga2022linkbert, shen2021topic, xie2021graph}. The recently
proposed CS-TAG benchmark dataset \cite{TAG_rethinking} with rich
raw text aims to facilitate research in TAG analysis. However, these
datasets oversimplify the evolving interactions in the real world by
representing them as static graphs, ignoring the inherent temporal
information present in real-world TAG scenarios, such as citation
networks with timestamped publications and social networks with
chronological user interactions. Consequently, there is an urgent
need for benchmark datasets that can accurately capture both dynamic
graph structures and rich text attributes of DyTAGs.

\xhdr{Proposed Work} To address this gap, we introduce a
comprehensive benchmark DTGB, which comprises eight large-scale
DyTAGs sourced from diverse domains including e-commerce, social
networks, multi-round dialogue, and knowledge graphs. Nodes and
edges in the DTGB dataset are associated with rich text descriptions
and edges are annotated with meaningful categories. The dataset
construction involves a meticulous process, including the selection
of data sources, the construction of the graph structure, and the
extraction of text and category information (detailed in
Section~\ref{sec:dataset_detail}). Compared with existing datasets
\cite{TAG_rethinking, madan2011sensing, kumar2019predicting,
huang2022dgraph}, which either lack raw text and temporal
annotations or are small in scale and devoid of long-term dynamic
structures, DTGB distinguishes itself with its \textit{rich text},
\textit{long-range historical information}, and \textit{large-scale
dynamic structures}, ensuring diverse and representative samples of
real-world DyTAG scenarios.

\begin{figure}[!t]
  \centering
  \includegraphics[width=0.95\linewidth]{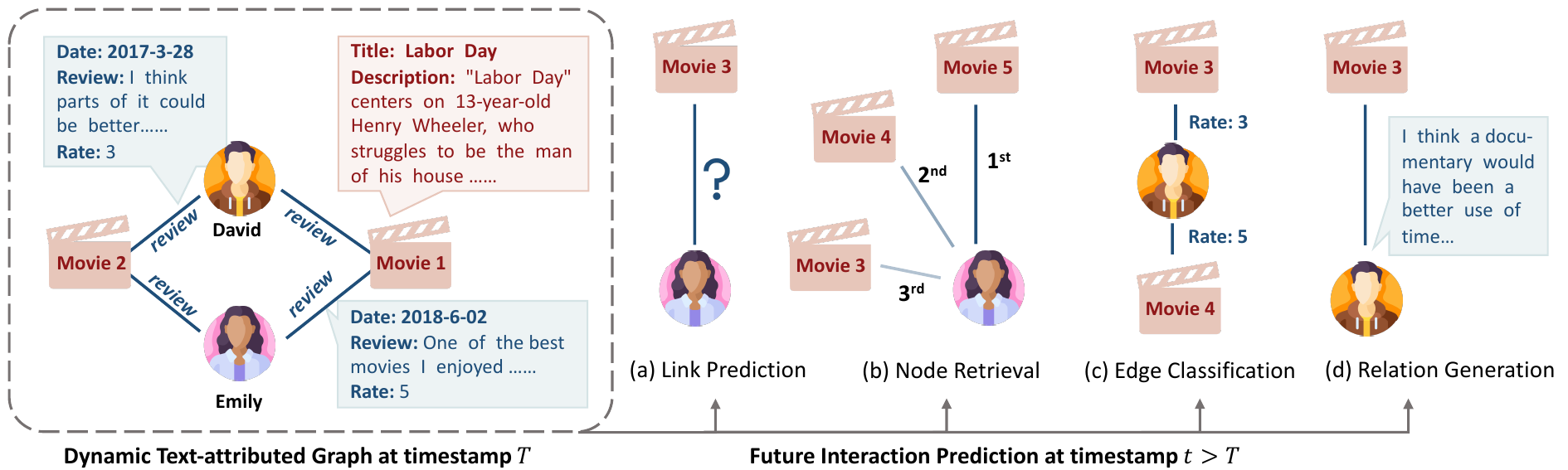}
  \caption{Dynamic text-attributed graph and evaluation tasks: a case study with movie reviews. Given a DyTAG with interactions before timestamp $T$, the tasks are to forecast future interactions in $t>T$, as well as their detailed interaction types and textual descriptions.}
  \label{fig:cover}
\end{figure}

With DTGB, we design four critical downstream evaluation tasks based
on real-world use cases, as shown in Figure~\ref{fig:cover}. Except
for the widely studied future link prediction task, we delve deeper
into three more interesting and challenging tasks: destination node
retrieval, edge classification, and textural relation generation,
which are neglected by previous works. These tasks are fundamental
to real-world applications that require the model to handle the
temporal evolution of both graph structure and textual information.
Our benchmarking results on 7 dynamic graph learning algorithms and
6 LLMs highlight these challenges and showcase the performance and
limitations of current algorithms (\eg the scalability issue of
memory-based models, neglect of the edge modeling, and weakness in
capturing long-range and semantic relevance), offering valuable
insights into integrating structural and textual dynamics. Our
\textbf{contributions} are summarized as follows:
\begin{compactitem}
    \item \textbf{First DyTAG Benchmark.} To the best of our knowledge, DTGB
is the first open benchmark specifically designed for dynamic
text-attributed graphs. We collect eight DyTAG datasets from a wide
range of domains and organize them in a unified structure, providing
a comprehensive testbed for model evaluation in this area.
    \item \textbf{Standardized Evaluation Protocol.} We design four critical
downstream tasks and standardize the evaluation process with DTGB.
This comprehensive evaluation highlights the unique challenges posed
by DyTAGs and demonstrates the utility of the dataset for assessing
algorithm performance.
    \item \textbf{Empirical Observation.} Our experimental results demonstrate
that rich textural information consistently enhances downstream
graph learning, such as destination node retrieval and edge
classification. This contributes to a deeper understanding of the
complexities involved in DyTAGs and offers guidance for future
research in this field.
\end{compactitem}

\section{Related Works}

\xhdr{Dynamic Graph Learning} Deep learning on dynamic graphs has
gained significant attention in various domains such as social
networks, transportation systems, and biological networks
\cite{dynamic_graph_survey, dynamic_network_survey,
dynamic_embedding, representation_dynamic_survey,
predicting_dynamic}. Previous works have proposed a few real-world
datasets \cite{DGB, TGB, TransactionTempGraph, shetty2004enron,
you2022roland}, which offer a comprehensive collection of temporal
interaction data across different contexts. Benchmark frameworks
\cite{benchtemp, pytorch_temp, DyGLib, robust_comparative}, such as
DyGLib \cite{DyGLib}, have been instrumental in standardizing the
evaluation of dynamic graph models, offering robust metrics for
assessing downstream performance. Recent advancements in temporal
graph models have significantly enhanced the ability to capture
time-evolving relationships in graph-structured data \cite{TGN,
TGAT, NAT, CAWN, Graphmixer, TCL, dyrep, nath2023tboost}, leading to
state-of-the-art performance in various tasks such as dynamic link
prediction and node classification on temporal graphs. However,
existing temporal graph datasets may lack node or edge attributes,
or contain simple node/edge features based on bag-of-words
\cite{BOW} or word2vec \cite{Word2vec} algorithms derived from the
associated text, which are limited in capturing the complicated
semantics of the text. In this work, we focus on dynamic graphs
where nodes and edges are associated with raw textual descriptions,
enabling richer, context-aware representations and more
sophisticated downstream tasks.

\xhdr{Text-attributed Graph Learning} Text-attributed graphs (TAG)
are widely used in various real-world applications. For example, in
citation networks, the text associated with each article provides
valuable information such as abstracts and titles
\cite{sen2008collective, Microsoft, yang2021graphformers}. CS-TAG
\cite{TAG_rethinking} offers standardized datasets with raw text,
facilitating research and methodological advances in TAG analysis.
Recently, large language models (LLMs) have demonstrated remarkable
capabilities in enhancing feature encoding and node classification
on TAGs \cite{empowerTAG, harnessing, pan2024distilling}. By
flattening graph structures and associated textual information into
prompts \cite{tang2024graphgpt, ye2024natural, zhao2023graphtext},
LLMs can leverage their strong language understanding and generation
abilities to improve TAG analysis tasks. However, all these TAGs
eliminate the temporal information within the graphs, which is
inherent and crucial in real-world scenarios. There has been limited
exploration of temporal relations in TAGs, which represents a missed
opportunity to evaluate the temporal awareness and reasoning ability
of LLMs.

\section{Dataset Details}
\label{sec:dataset_detail}

\xhdr{Motivation of DTGB} To investigate the necessity of a
comprehensive benchmark dataset for dynamic text-attributed graphs,
we first survey various dynamic graphs and text-attributed graphs
previously utilized in the literature. We observed that most
commonly used dynamic graphs are essentially text-attributed.
Simultaneously, many popular text-attributed graphs have inherent
temporal information. For instance, the well-known dynamic graph
dataset \texttt{tgbl-review} \cite{TGB} and the commonly used TAG
dataset \texttt{Books-Children} \cite{TAG_rethinking} are both
derived from the Amazon product review network
\cite{ni2019justifying} which is intrinsically associated with both
the text attributes of users and products and the time annotations
of user-item interactions. However, \texttt{tgbl-review} only
contains numerical attributes derived from the raw text and
\texttt{Books-Children} ignores all temporal information, which
highly limits the full exploration of the performance for downstream
tasks.

While these previous datasets are frequently used, they possess
obvious inadequacies when exploring the effectiveness of dynamic
graph learning methods in handling real-world scenarios. Firstly,
existing dynamic graph datasets lack the availability of raw textual
information, bringing challenges to investigating the benefits of
text attribute modeling on real-world applications. Secondly, most
existing dynamic graph datasets lack reasonable temporal
segmentation and aggregation, making their edges distribution quite
sparse in the time dimension (\eg \texttt{MOOC}
\cite{kumar2019predicting} and \texttt{tgbl-wiki} \cite{TGB}). This
brings challenges to investigating the structure dependency and
evolution for dynamic graphs. Lastly, although TAGs are enriched
with node text attributes, they tend to miss edge text and time
annotations, making them fail to faithfully reflect the challenges
in modeling real-world scenarios.

\xhdr{Dataset Construction} To address these challenges, we collect
resources from different domains and follow a rigorous process to
construct the comprehensive benchmark dataset DTGB for DyTAGs. We
carefully select data sources from various domains to ensure
diversity and relevance. For graph construction, the redundant and
low-quality records are first filtered out and we divide the data
into discrete time intervals. In each time interval, nodes and edges
are flexibly identified for different domains. Nodes could represent
users, products, questions, \textit{etc.} while edges represent
relationships such as transactions and reviews. For text and
category extraction, we organize multiple text descriptions from the
source data and remove the meaningless or garbled characters and
low-quality text. We categorize edges based on predefined criteria
relevant to real-world use cases such as product ratings and content
topics. This process ensures that the dataset accurately reflects
the complexity of real-world DyTAG. Taking \texttt{Googlemap CT} and
\texttt{Amazon movies} as an example, which are extracted from
Recommender Systems and Personalization Datasets
\cite{lakkaraju2013s}. Nodes represent users or items, while edges
indicate review relation between users and items. The original data
is first reduced to a $k$-core subgraph, indicating that each user
or item has at least $k$ reviews. Edges are segmented by days and
edges within each day are aggregated as a subgraph. Edge categories
are integers from 1 to 5, derived from the ratings from users to
items. Node text includes the basic information of the item (\eg
name, description, and category). Edge text is the raw review from
users. Detailed descriptions of all the datasets can be found in
Appendix~\ref{appd:dataset_intro}.

\begin{table}[t] \small
\caption{Statistics of datasets and comparison with existing
datasets.} \centering \scalebox{0.7}{
  \begin{tabular}{c||c|ccccccc}
    \toprule
    &\textbf{Dataset} & Nodes & Edges & Edge Categories & Timestamps & Domain & Text Attributes & Bipartite Graph \\
    \midrule
    \multirow{5}{*}{\begin{tabular}[c]{@{}c@{}}Previous\\ Dynamic \\ Graphs\end{tabular}} &\textbf{tgbn-trade} &255  &468,245 &N.A. &32 &Trade &\xmark &\xmark \\
    &\textbf{tgbl-wiki} &9,227 &157,474 &N.A. &152,757 &Interaction &\xmark &\xmark \\
    &\textbf{tgbl-review} &352,637 &4,873,540 &N.A. &6,865 &E-commerce &\xmark &\cmark \\
    &\textbf{MOOC} &7,144 &411,749 &N.A. &345,600 &Interaction &\xmark &\cmark \\
    &\textbf{LastFM} &1,980 &1,293,103 &N.A. &1,283,614 &Interaction &\xmark & \cmark\\
\midrule[0.3pt]\midrule[0.3pt]
    \multirow{5}{*}{\begin{tabular}[c]{@{}c@{}}Previous\\ TAGs \end{tabular}}  &\textbf{ogbn-arxiv-TA} &169,343 & 1,166,243 &N.A. &N.A. &Academic &Node &\xmark \\
    &\textbf{CitationV8} &1,106,759 &6,120,897 &N.A. &N.A. &Academic &Node &\xmark \\
    &\textbf{Books-Children} &76,875 &1,554,578 &N.A. &N.A. &E-commerce &Node &\xmark \\
    &\textbf{Ele-Computers} &87,229 & 721,081 &N.A. &N.A. &E-commerce &Node &\xmark \\
    &\textbf{Sports-Fitness} &173,055 &1,773,500 &N.A. &N.A. &E-commerce &Node &\xmark \\
\midrule[0.3pt]\midrule[0.3pt]
    \multirow{8}{*}{Ours}  &\textbf{Enron} &42,711 &797,907 &10 &1,006 & E-mail &Node \& Edge & \xmark \\
    &\textbf{GDELT} &6,786 &1,339,245 &237 &2,591 & Knowledge graph & Node \& Edge & \xmark \\
    &\textbf{ICEWS1819} &31,796 &1,100,071 &266 &730 & Knowledge graph &Node \& Edge & \xmark \\
    &\textbf{Stack elec} &397,702 &1,262,225 &2 &5,224 & Multi-round dialogue &Node \& Edge & \cmark \\
    &\textbf{Stack ubuntu} &674,248 &1,497,006 &2 &4,972 & Multi-round dialogue &Node \& Edge & \cmark \\
    &\textbf{Googlemap CT} &111,168 &1,380,623 &5 &55,521 & E-commerce &Node \& Edge & \cmark \\
    &\textbf{Amazon movies} &293,566 &3,217,324 &5 &7,287 & E-commerce &Node \& Edge & \cmark \\
    &\textbf{Yelp} &2,138,242 &6,990,189 &5 &6,036 & E-commerce  &Node \& Edge & \cmark \\
\bottomrule
  \end{tabular}}
  \label{tab:datasets}
\end{table}

\begin{figure}[t]
  \centering
  \includegraphics[width=0.9\linewidth]{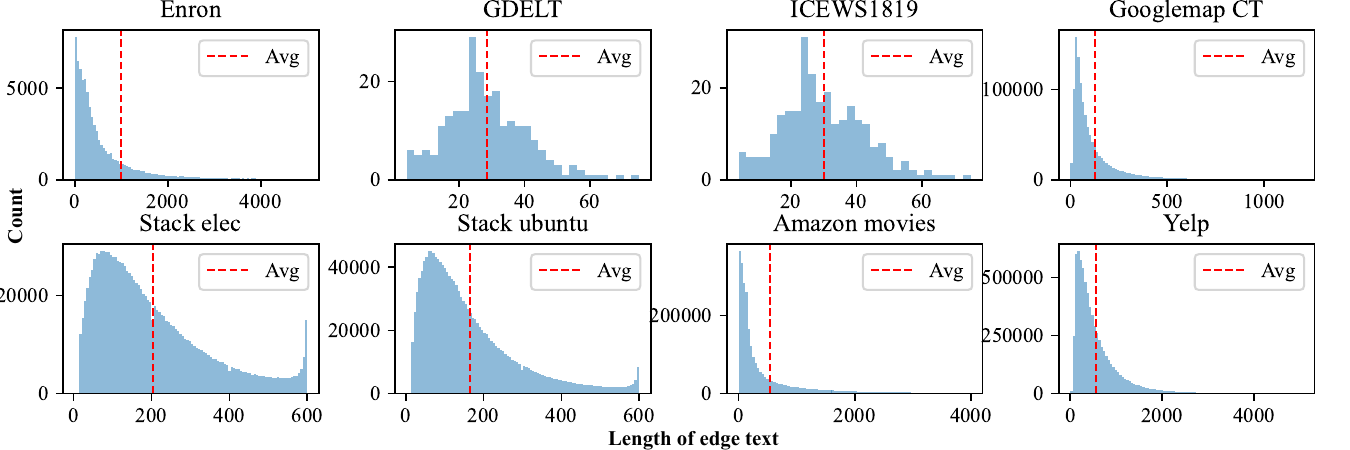}
  \caption{Distribution of edge text lengths on the DTGB datasets.}
  \label{fig:text_length}
\end{figure}

\begin{figure}[t]
  \centering
  \includegraphics[width=0.9\linewidth]{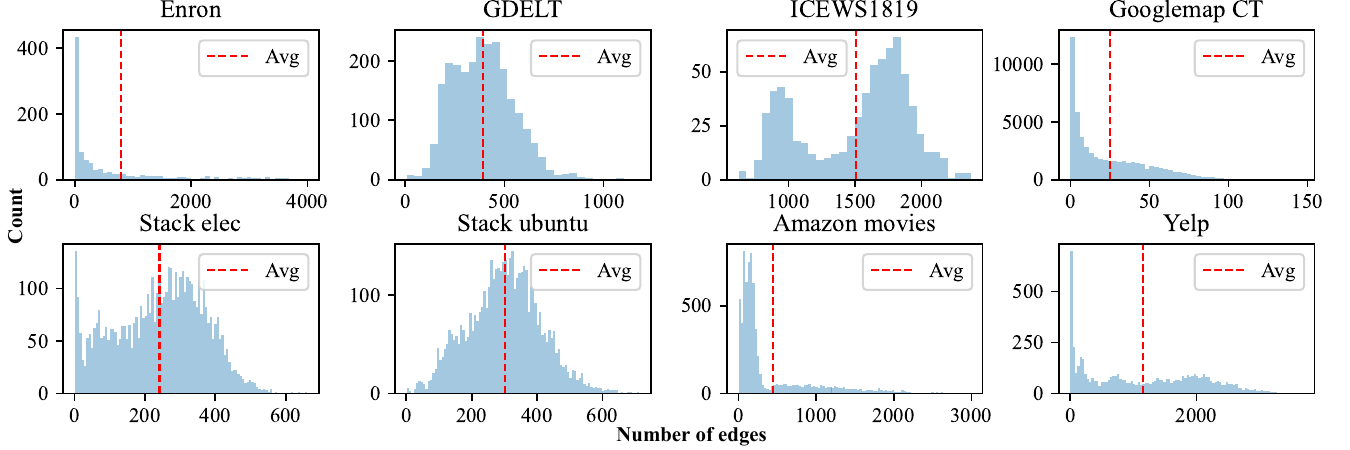}
  \caption{Distribution of the numbers of edges in each timestamp on the DTGB datasets.}
  \label{fig:time_edge}
\end{figure}

\xhdr{Distribution and Statistics} Table~\ref{tab:datasets} gives
the statistics of previous datasets and DTGB datasets. One can
notice that compared with previous dynamic graph datasets, our
datasets are characterized by edge categories and text attributes at
both node and edge levels. Our dataset includes small, medium, and
large graphs with various distributions from four different domains,
encompassing both bipartite and non-bipartite, long-range and
short-range dynamic graphs. We further study the data distribution
to better understand our benchmark datasets. As shown in
Figure~\ref{fig:text_length} and Figure~\ref{fig:time_edge},
datasets from the same domain exhibit similar distributions. For
example, knowledge graph datasets \texttt{GDELT} and
\texttt{ICEWS1819} both approximate Gaussian distributions in edge
text length, and e-commerce datasets \texttt{Googlemap CT},
\texttt{Amazon movies}, and \texttt{Yelp} show long-tail
distributions in the number of edges per timestamp. This
demonstrates that our datasets have faithfully preserved the
characteristics of data from different domains. More detailed
analysis of our datasets can be found in
Appendix~\ref{appd:dataset_analysis}.

\section{Formulation and Benchmarking Tasks on DyTAG}

\xhdr{DyTAG Formulation} A DyTAG can be defined as $\mathcal{G} = \{
\mathcal{V}, \mathcal{E}\}$, where $\mathcal{V}$ is the node set,
$\mathcal{E} \subset \mathcal{V} \times \mathcal{V}$ is the edge
set. Let $\mathcal{T}$ denote the set of observed timestamps,
$\mathcal{D}$, $\mathcal{R}$ and $\mathcal{L}$ are the set of node
text descriptions, edge text descriptions, and edge categories,
respectively. Each $v \in \mathcal{V}$ is associated with a text
description $d_{v} \in \mathcal{D}$. Each $(u,v) \in \mathcal{E}$
can be represented as $(r_{u,v},l_{u,v},t_{u,v})$ with a text
description $r_{u,v} \in \mathcal{R}$, a category $l_{u,v} \in
\mathcal{L}$ and a timestamp $t_{u,v} \in \mathcal{T}$ to indicate
the occurring time of this edge. We use
$\mathcal{G}_T=\{\mathcal{V}_T,\mathcal{E}_T\}$ to represent the
DyTAG containing interactions occurred before timestamp $T$. We
summarize the important notations used in
Appendix~\ref{appd:notion}.

\xhdr{Future Link Prediction} Future link prediction is commonly
used in previous literature \cite{TGB, DyGLib, DGB} to evaluate the
performance of dynamic graph learning methods, which aims to predict
whether two nodes will be linked in the future given the history
edges. In dynamic text-attributed graphs, the new linkage between
nodes not only depends on their static semantics brought by node
text but also on the interaction context brought by edge text
semantics. The future link prediction task in the DyTAG setting can
be viewed as the simplification of many real-world applications, \eg
predicting whether one person will e-mail another based on the
content of their history e-mails. Formally, given the DyTAG
$\mathcal{G}_T$ which contains edges before timestamp $T$, future
link prediction aims to predict whether an interaction will happen
between nodes $u$ and $v$ at timestamp $T+1$. In the inductive
setting, either $u$ or $v$ are new nodes not contained in
$\mathcal{V}_T$.

\xhdr{Destination Node Retrieval} While future link prediction has
been widely used, this task has several limitations for a reliable
evaluation. Its performance largely depends on the quality and the
size of the sampled negative samples, and the binary classification
metrics are sensitive to the model fluctuations. Destination node
retrieval is a novel task that aims to rank the most likely interact
nodes for a given node based on its interaction history. This
approach is more stable as it considers the relative relevance among
the entire node set and is more applicable to real-world scenarios
(\eg personalized recommendation). Formally, given node $u$ and
$\mathcal{G}_T$, node retrieval aims to rank the nodes in
$\mathcal{V}_T$ based on their possibilities of interacting with $u$
in timestamp $T+1$. In the inductive setting, $u$ is a new node not
contained in $\mathcal{V}_T$.

\xhdr{Edge Classification} Edge classification is an essential
evaluation task for DyTAG, which is under-explored by previous
dynamic graph benchmarks \cite{TGB, DyGLib, DGB}. The dynamic graph
learning models leverage both rich textual information and
historical interactions to predict the categories of relations
between them (\eg the review rating in the future). Formally, given
a DyTAG $\mathcal{G}_T$ with edges up to timestamp $T$, edge
classification aims to predict the category of a potential edge at
timestamp $T+1$, utilizing both node and edge textual attributes and
historical interactions.

\xhdr{Textural Relation Generation} Textural relation generation is
a novel task that seeks to leverage historical interactions and
their associated text to generate future relation context. While
previous studies on TAGs \cite{TAG_rethinking, harnessing} have
mainly focused on graph structure learning, such as predicting new
edges or node labels, generating the actual textual content for
future edges remains an under-explored challenge. To address this
gap, large language models (LLMs) are employed as backbones, due to
their powerful capability in understanding and generating natural
language. Specifically, given two nodes $u$ and $v$ for which we aim
to predict their future textual interaction, we provide the LLM with
their node descriptions as well as the historical one-hop
interactions involving either $u$ or $v$. We then prompt the LLM to
generate the predicted interaction text in an autoregressive manner.
This task not only serves as a challenging benchmark for evaluating
LLMs' ability to understand the co-evolution of graph structures and
natural language, but also holds promise for enhancing LLMs'
representations by incorporating the inductive biases of structured
data during pretraining in the future.

\section{Experiments}
\label{sec:experiment}

\xhdr{Baselines} (1) For the future link prediction, destination
node retrieval, and edge classification tasks, we use the dynamic
graph learning models as the baselines. We evaluate 7 popular and
state-of-the-art models: JODIE \cite{JODIE}, DyRep \cite{dyrep},
TGAT \cite{TGAT}, CAWN \cite{CAWN}, TCL \cite{TCL}, GraphMixer
\cite{Graphmixer} and DyGFormer \cite{DyGLib}. (2) For the textual
relation generation task, we benchmark four open-source large
language models: Mistral 7B \cite{jiang2023mistral}, Vicuna 7B/13B
\cite{vicuna}, Llama-3 8B \cite{llama3}, and two closed-source large
language models GPT3.5-turbo and the most recent GPT4o through API
service. Refer to Appendix~\ref{appd:baseline} for more details.

\xhdr{Evaluation Metrics} For the edge classification task, we use
Weighted Precision, Weighted Recall, and Weighted F1 score to
evaluate the model performance. For the future link prediction task,
we follow previous works \cite{TGB, DyGLib} and adopt the Average
Precision (AP) and Area Under the Receiver Operating Characteristic
Curve (AUC-ROC) as the evaluation metrics. For the node retrieval
task, we use Hits@$k$ as the metric, which reports whether the
correct item appears within the top-$k$ results generated by the
model. To evaluate the generated textual relation, we use BERTScore
\cite{bertscore}, which leverages a pre-trained language model and
calculates the cosine similarity between the prediction and ground
truth. The detailed definitions are provided in
Appendix~\ref{appd:metirc}.

\xhdr{Experimental Settings} For dynamic graph learning models, we
follow the implementations from
DyGLib\footnote{\url{https://github.com/yule-BUAA/DyGLib}}
\cite{DyGLib}. All data loading, training, and evaluation processes
are performed uniformly, following DyGLib. To integrate the textual
information, we use the Bert-base-uncased model \cite{Bert} to
encode the node and edge texts as the initialization of the node and
edge representations. We chronologically split each dataset into
train/validation/test sets by 7:1.5:1.5. For the textual relation
generation task, open-source LLMs are implemented with Huggingface
\cite{HF}. We also use the parameter-efficient fine-tuning method,
LoRA \cite{LORA}, to fine-tune the $\mathbf{Q}$, $\mathbf{K}$,
$\mathbf{V}$, and $\mathbf{O}$ matrices within LLM for better text
generation. We run all the models five times with different seeds
and report the average performance to eliminate deviations.
Experiments are conducted on NVIDIA A40 with 48 GB memory.

\xhdr{Implementation Details of Dynamic Graph Models} For all of the
edge classification task, future link prediction task, and
destination node retrieval task, we use Adam \cite{kingma2014adam}
for optimization. All the models are trained for 500 epochs and use
the early stopping strategy with a patience of 5. The batch size is
set as 256. For the edge classification task, after obtaining the
representations of the target node and source node, we feed the
concatenated representations of two nodes into a multi-layer
perceptron to perform the multi-class classification. We employ the
cross-entropy loss function to supervise the training in this task.
The future link prediction task and the destination node retrieval
task share the same training process where a multi-layer perception
is used to obtain the possibility scores and the binary
cross-entropy loss function is used for supervision. After obtaining
the possibility scores of test samples, we traverse different
thresholds to get the AP and AUC-ROC metrics for the future link
prediction task, and rank the possibility scores of all candidates
to get the Hits@k metric for the destination node retrieval task. We
perform the grid search based on the performance of the validation
set to find the best settings of some critical hyperparameters,
where the searched ranges and related models are shown in
Table~\ref{tab:hyperparameter}. We use the vanilla recurrent neural
network as the memory updater of JODIE and DyRep. For CAWN, the time
scaling factor is set as $1e-6$, and the length of each walk is set
as 2. To integrate the text information, we use the pre-trained
language model (\ie
Bert-base-uncased\footnote{\url{https://huggingface.co/google-bert/bert-base-uncased}})
to get the representations of text attributes, and then these
representations are used to initialize the embeddings of nodes and
edges in models. The dimensions of the pre-trained representations
are 768 and the maximum text length is set as 512.

\begin{table}[t] \small
 \centering
  \caption{Searched ranges of hyperparameters and the related dynamic graph learning models.}
  \label{tab:hyperparameter}
  \resizebox{0.85\linewidth}{!}{
  \begin{tabular}{c|cc}
    \toprule
    Hyperparameters &Searched Ranges & Related Models\\
    \midrule
    Dropout Rate &[0.0, 0.2, 0.4, 0.6] & All of 7 models \\
    Sampling Size &[10, 20, 30] & DyRep, TGAT, TCL, GraphMixer \\
    Learning Rate &[0.0001, 0.0005, 0.001] & All of 7 models \\
    Sampling Strategies &[uniform,recent] &TCL, GraphMixer \\
    Number of Walks &[16, 32, 64, 128] &CAWN \\
    Sequence Length &[32, 64, 128, 256, 512, 1024, 2048, 4096] &DyGFormer \\
    Patch Size &[1, 2, 4, 8, 16, 32, 64, 128] &DyGFormer \\
    Number of GNN Layers &[1, 2, 3] &DyRep, TGAT \\
    Number of Transformer Layers &[1, 2, 3] &TCL, DyGFormer \\
    Number of Attention Heads &[2, 4, 6, 8] &DyRep, TGAT, CAWN, TCL, DyGFormer \\
    \bottomrule
  \end{tabular}}
\end{table}

\xhdr{Implementation Details of Large Language Models} For the
textual relation generation task, the inputs to LLMs include the
text attribute of the source node and target node, the recent $k$
edges from the source node and the corresponding text, and the
recent $k$ edges from the target node and the corresponding text.
The detailed description of prompts can be found in
Appendix~\ref{appd:prompt} and the experimental results of using
different history length can be found in
Appendix~\ref{appd:relation_generation_result}. During inference, we
set the temperature as 0.7 and the nucleus sampling size (\ie
$top\_p$) is set as 0.95. These two hyperparameters are used to
control the randomness of LLMs' output. The repetition penalty is
set as 1.15 to discourage the repetitive and redundant output. The
maximum number of tokens that the LLM generates is set as 1024.
During fine-tuning, the LLMs are loaded in 8-bits and the rank of
LoRA is set as 8. We set the batch size as 2 with only one epoch and
the gradient accumulation step is set as 8. The learning rate is set
as 0.0002 and we use AdamW \cite{loshchilov2017decoupled} for
optimization. The scaling hyperparameter $lora\_alpha$ is set as 32
and the dropout rate during fine-tuning is set as 0.05.

\subsection{Edge Classification}

\begin{table}[t]
\caption{Performance of dynamic graph learning methods in the edge
classification task when using Bert-encoded embeddings to initialize
the model representations. OOM means out-of-memory.} \centering
\resizebox{1\textwidth}{!}{
\begin{tabular}{c|c|ccccccc}
\toprule
\textbf{Datasets}
&\textbf{Models}
&\multicolumn{1}{c}{\textbf{JODIE}}
&\multicolumn{1}{c}{\textbf{DyRep}}
&\multicolumn{1}{c}{\textbf{TGAT}}
&\multicolumn{1}{c}{\textbf{CAWN}}
&\multicolumn{1}{c}{\textbf{TCL}}
&\multicolumn{1}{c}{\textbf{GraphMixer}}
&\multicolumn{1}{c}{\textbf{DyGFormer}}\\
\midrule
&Precision &0.6568 $\pm$ 0.0043 &\textbf{0.6635 $\pm$ 0.0052} &0.6148 $\pm$ 0.0012 &0.6076 $\pm$ 0.0070 &0.5530 $\pm$ 0.0079 &0.6313 $\pm$ 0.0024 &0.6601 $\pm$ 0.0067\\
 \textbf{Enron} &Recall &\textbf{0.6472 $\pm$ 0.0039}  &0.6390 $\pm$ 0.0089 &0.5530 $\pm$ 0.0001 &0.5783 $\pm$ 0.0094 &0.5394 $\pm$ 0.0061 &0.5735 $\pm$ 0.0015 &0.5802 $\pm$ 0.0071\\
 &F1 &\textbf{0.6478 $\pm$ 0.0065} &0.6432 $\pm$ 0.0062   &0.5519 $\pm$ 0.0028   &0.5685 $\pm$ 0.0132   &0.5177 $\pm$ 0.0044   &0.5507 $\pm$ 0.0019   &0.5604 $\pm$ 0.0063\\

\midrule

 &Precision &0.1361 $\pm$ 0.0036   &0.1451 $\pm$ 0.0071 &0.1241 $\pm$ 0.0056 &\textbf{0.1781 $\pm$ 0.0011}  &0.1229 $\pm$ 0.0021   &0.1293 $\pm$ 0.0026 &0.1775 $\pm$ 0.0041\\
 \textbf{GDELT} &Recall &0.1338 $\pm$ 0.0013   &0.1365  $\pm$ 0.0013  &0.1321 $\pm$ 0.0012   &0.1545 $\pm$ 0.0001   &0.1235 $\pm$ 0.0047   &0.1320 $\pm$ 0.0008   &\textbf{0.1580 $\pm$ 0.0052}\\
 &F1 &0.0992 $\pm$ 0.0009  &0.1039  $\pm$ 0.0012  &0.0967 $\pm$ 0.0010   &\textbf{0.1340 $\pm$ 0.0012}  &0.0987 $\pm$ 0.0051   &0.1014 $\pm$ 0.0017   &0.1291 $\pm$ 0.0068\\

\midrule

 &Precision &0.3106 $\pm$ 0.0023   &0.3270 $\pm$ 0.0025 &0.3013 $\pm$ 0.0007 &\textbf{0.3451 $\pm$ 0.0023}  &0.3212 $\pm$ 0.0096   &0.2999 $\pm$ 0.0022 &0.3297 $\pm$ 0.0034\\
 \textbf{ICEWS1819} &Recall &0.3494 $\pm$ 0.0018   &0.3636 $\pm$ 0.0020 &0.3512 $\pm$ 0.0006 &\textbf{0.3676 $\pm$ 0.0034}  &0.3517 $\pm$ 0.0009   &0.3502 $\pm$ 0.0001   &0.3632 $\pm$ 0.0026\\
 &F1 &0.2965 $\pm$ 0.0008  &0.3097 $\pm$ 0.0006   &0.2908 $\pm$ 0.0008   &\textbf{0.3156 $\pm$ 0.0057}  &0.2939 $\pm$ 0.0022   &0.2903 $\pm$ 0.0008   &0.3079 $\pm$ 0.0027\\

\midrule

 &Precision &0.6163 $\pm$ 0.0032   &0.6073 $\pm$ 0.0019 &0.6160 $\pm$ 0.0001 &0.6166 $\pm$ 0.0023   &\textbf{0.6213 $\pm$ 0.0087}  &0.6171 $\pm$ 0.0020 &0.6166 $\pm$ 0.0003\\
 \textbf{Googlemap CT} &Recall &0.6871 $\pm$ 0.0002    &0.6827 $\pm$ 0.0006 &0.6862 $\pm$ 0.0002 &0.6870 $\pm$ 0.0001   &0.6875 $\pm$ 0.0001   &0.6872 $\pm$ 0.0003   &\textbf{0.6877 $\pm$ 0.0002}\\
 &F1 &0.6189 $\pm$ 0.0016  &0.6134 $\pm$ 0.0006   &0.6225 $\pm$ 0.0015   &0.6187 $\pm$ 0.0003   &\textbf{0.6230 $\pm$ 0.0003}  &0.6185 $\pm$ 0.0005   &0.6196 $\pm$ 0.0008\\

 \midrule

 &Precision &OOM    &OOM &0.6265 $\pm$ 0.0046  &0.6167 $\pm$ 0.0094   &\textbf{0.6325 $\pm$ 0.0023}  &0.6074 $\pm$ 0.0039   &0.6026 $\pm$ 0.0471\\
 \textbf{Stack elec} &Recall &OOM &OOM  &0.7205 $\pm$ 0.0094   &0.6313 $\pm$ 0.0462   &\textbf{0.7474 $\pm$ 0.0004}  &0.7412 $\pm$ 0.0061   &0.5891 $\pm$ 0.2747\\
 &F1 &OOM &OOM  &\textbf{0.6496 $\pm$ 0.0032}  &0.6209 $\pm$ 0.0216   &0.6420 $\pm$ 0.0003   &0.6412 $\pm$ 0.0005   &0.4860 $\pm$ 0.2686\\

 \midrule

 &Precision &OOM    &OOM &0.6858 $\pm$ 0.0047  &0.6921 $\pm$ 0.0040   &0.6915 $\pm$ 0.0118   &\textbf{0.6930 $\pm$ 0.0028}  &0.6789 $\pm$ 0.0490\\
\textbf{Stack ubuntu} &Recall &OOM  &OOM    &\textbf{0.7921 $\pm$ 0.0012}  &0.5650 $\pm$ 0.1015   &0.7880 $\pm$ 0.0026   &0.7902 $\pm$ 0.0130   &0.7494 $\pm$ 0.0991\\
 &F1 &OOM &OOM  &0.7201 $\pm$ 0.0013   &0.6002 $\pm$ 0.0738   &\textbf{0.7219 $\pm$ 0.0046}  &0.7214 $\pm$ 0.0014   &0.7033 $\pm$ 0.0294\\

 \midrule

 &Precision &0.5923 $\pm$ 0.0046   &OOM &0.5878 $\pm$ 0.0037  &0.5931 $\pm$ 0.0021   &0.5861 $\pm$ 0.0017   &0.5934 $\pm$ 0.0010   &\textbf{0.5943 $\pm$ 0.0028}\\
\textbf{Amazon movies} &Recall &0.6713 $\pm$ 0.0038     &OOM   &0.6711 $\pm$ 0.0043 &0.6731 $\pm$ 0.0012 &0.6692 $\pm$ 0.0051   &0.6720 $\pm$ 0.0084   &\textbf{0.6737 $\pm$ 0.0058}\\
 &F1 &0.6042 $\pm$ 0.0072 &OOM &0.5917 $\pm$ 0.0051   &0.5965 $\pm$ 0.0018   &0.5814 $\pm$ 0.0037   &0.5991 $\pm$ 0.0064   &\textbf{0.6050 $\pm$ 0.0084}\\

\midrule

 &Precision &OOM    &OOM &0.6186 $\pm$ 0.0018  &0.6162 $\pm$ 0.0017   &0.6255 $\pm$ 0.0026   &0.6295 $\pm$ 0.0085   &\textbf{0.6407 $\pm$ 0.0016}\\
\textbf{Yelp} &Recall &OOM  &OOM    &0.6686 $\pm$ 0.0072   &0.6654 $\pm$ 0.0038   &0.6735 $\pm$ 0.0038   &0.6736 $\pm$ 0.0108   &\textbf{0.6812 $\pm$ 0.0072}\\
 &F1 &OOM   &OOM    &0.6219 $\pm$ 0.0103   &0.6085 $\pm$ 0.0024   &0.6247 $\pm$ 0.0084   &0.6234 $\pm$ 0.0174   &\textbf{0.6359 $\pm$ 0.0057}\\

\bottomrule
\end{tabular}}
\label{tab:edge_classification}
\end{table}

\begin{figure}[t]
\centering \subfigure[Googlemap
CT]{\includegraphics[width=0.32\linewidth]{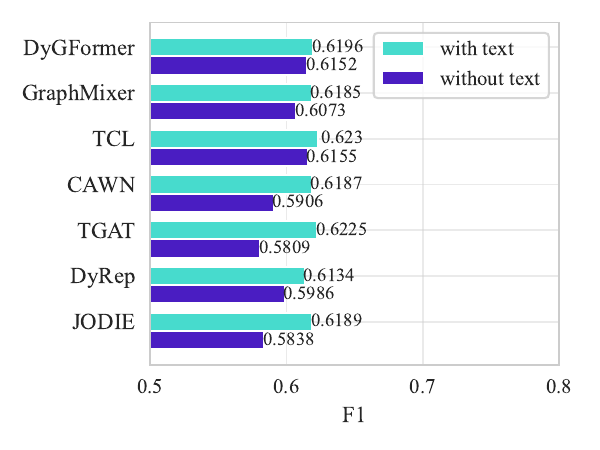}}
\subfigure[ICEWS1819]{\includegraphics[width=0.32\linewidth]{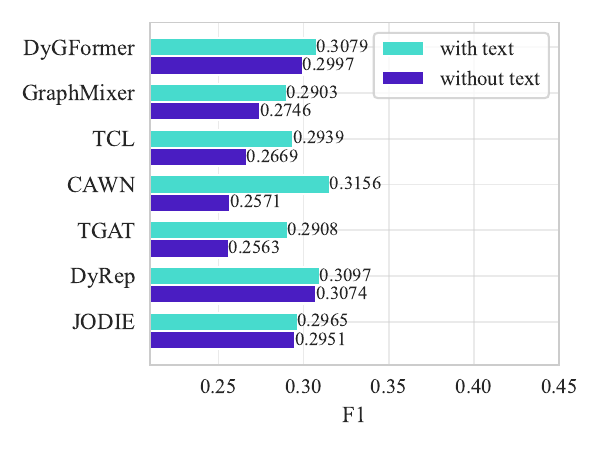}}
\subfigure[Enron]{\includegraphics[width=0.32\linewidth]{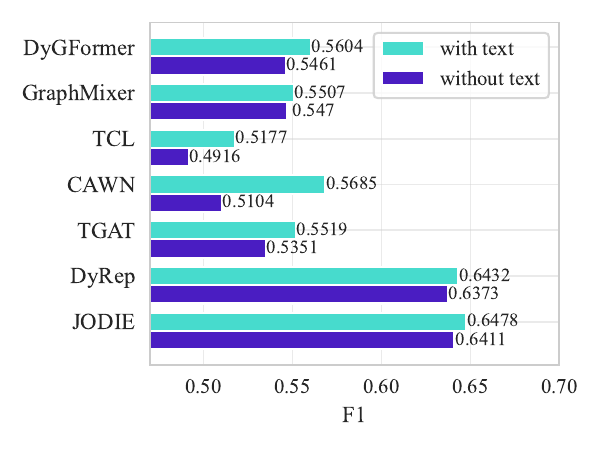}}
\caption{Edge classification performance with and without text
attributes.} \label{fig:edge_classification_bert_ablation}
\end{figure}

Following previous work \cite{DyGLib}, we use a multi-layer
perceptron to take the concatenated representations of two nodes as
inputs and return the probabilities of the edge categories. The
performance of different models with Bert initialization is shown in
Table~\ref{tab:edge_classification}, where the best results for each
dataset are shown in bold. We observe that existing models fail to
achieve satisfactory performance in this task, especially on
datasets with a large number of categories (\eg \texttt{GDELT} and
\texttt{ICEWS1819}). This can be attributed to the fact that these
models typically neglect edge information modeling in their
architectures, which is extremely important for edge classification
applications on DyTAGs. In
Figure~\ref{fig:edge_classification_bert_ablation}, we report the
model performance with and without text attributes. It demonstrates
that text information consistently helps models achieve better
performance on each dataset, verifying the necessity of integrating
text attributes into temporal graph modeling. Using Bert-encoded
embedding as initialization serves as a preliminary strategy for
dynamic textual modeling, lifting the future opportunities for more
advanced embedding. We provide the complete results for other
datasets in Appendix~\ref{appd:edge_classification_result}.

\subsection{Future Link Prediction}

We report the performance of different models in the AUC-ROC metric
for future link prediction under transductive and inductive settings
in Table~\ref{tab:link_prediction}. We have two main observations:
(1) Most models achieve better performance when using text
attributes. However, the performance of memory-based models (\ie
JODIE and DyRep) may be degraded. This decline occurs because these
models incrementally update the nodes' representations based on the
numerical attributes of edges. The Bert-encoded initialization of
edges will potentially mislead the update process. This observation
demonstrates the limitations of simply using pre-trained embeddings
to integrate the text information, showing the necessity of
proposing an advanced integration strategy that can adapt to
different models. (2) Larger performance improvements are observed
in the inductive setting. This is because the text attributes can
provide valuable information for new nodes that are difficult to
distinguish using existing methods. This observation shows the
effectiveness of integrating text information to handle zero-shot
dynamic graph problems (\eg cold-start in recommendation
\cite{vartak2017meta}). We provide the complete experimental results
of future link prediction in
Appendix~\ref{appd:link_prediction_result}.

\begin{table*}[t]
\caption{AUC-ROC for future link prediction. \textit{tr.} means
transductive setting and \textit{in.} means inductive setting.
\textbf{Text} means whether to use Bert-encoded embeddings for
initialization.} \centering \resizebox{1\textwidth}{!}{
\begin{tabular}{c|c|c|ccccccc}
\toprule
\textbf{}
&\textbf{Datasets}
&\textbf{Text}
&\multicolumn{1}{c}{\textbf{JODIE}}
&\multicolumn{1}{c}{\textbf{DyRep}}
&\multicolumn{1}{c}{\textbf{TGAT}}
&\multicolumn{1}{c}{\textbf{CAWN}}
&\multicolumn{1}{c}{\textbf{TCL}}
&\multicolumn{1}{c}{\textbf{GraphMixer}}
&\multicolumn{1}{c}{\textbf{DyGFormer}}\\
\midrule
&\multirow{2}{*}{\textbf{Enron}}  &\xmark &\textbf{0.9712 $\pm$ 0.0097} &0.9545 $\pm$ 0.0023 &0.9511 $\pm$ 0.0011 &0.9652 $\pm$ 0.0012 &0.9604 $\pm$ 0.0079 &0.9254 $\pm$ 0.0046 &0.9653 $\pm$ 0.0015\\

& &\cmark &0.9731 $\pm$ 0.0052 &0.9274 $\pm$ 0.0026 &0.9681 $\pm$ 0.0026 &0.9740 $\pm$ 0.0007 &0.9618 $\pm$ 0.0025 &0.9567 $\pm$ 0.0013 &\textbf{0.9779 $\pm$ 0.0014}\\

\cmidrule{2-10}

 &\multirow{2}{*}{\textbf{ICEWS1819}}&\xmark &0.9821 $\pm$ 0.0095 &0.9799 $\pm$ 0.0039 & 0.9787 $\pm$ 0.0065 &0.9815 $\pm$ 0.0041 &0.9842 $\pm$ 0.0036 &0.9399 $\pm$ 0.0079 &\textbf{0.9865 $\pm$ 0.0024}\\

 & &\cmark &0.9741 $\pm$ 0.0113 &0.9632 $\pm$ 0.0027 &0.9904 $\pm$ 0.0039 &0.9857  $\pm$ 0.0018 &\textbf{0.9923 $\pm$ 0.0012} &0.9863 $\pm$ 0.0024 &0.9888  $\pm$ 0.0015\\

 \cmidrule{2-10}

\textit{tr.} &\multirow{2}{*}{\textbf{Googlemap CT}} &\xmark &OOM &OOM &0.8537 $\pm$ 0.0153 &\textbf{0.8543 $\pm$ 0.0027} &0.7740 $\pm$ 0.0013 &0.7087 $\pm$ 0.0088 & 0.7864 $\pm$ 0.0047\\

 & &\cmark &OOM &OOM &\textbf{0.9049 $\pm$ 0.0071} &0.8687 $\pm$ 0.0063  &0.8348 $\pm$ 0.0094 &0.8095 $\pm$ 0.0014 &0.8207 $\pm$ 0.0018\\

\cmidrule{2-10}

&\multirow{2}{*}{\textbf{GDELT}} &\xmark &0.9562 $\pm$ 0.0027 &0.9477 $\pm$ 0.0011 &0.9341 $\pm$ 0.0046 &0.9419  $\pm$ 0.0026 &0.9571 $\pm$ 0.0007 &0.9316 $\pm$ 0.0021 &\textbf{0.9648 $\pm$ 0.0007}\\

 & &\cmark &0.9533 $\pm$ 0.0020 &0.9453 $\pm$ 0.0018 &0.9595 $\pm$ 0.0033 &0.9600 $\pm$ 0.0061 &0.9619  $\pm$ 0.0008 &0.9552 $\pm$ 0.0018 &\textbf{0.9662 $\pm$ 0.0003}\\

 \midrule

 &\multirow{2}{*}{\textbf{Enron}}&\xmark &0.8745 $\pm$ 0.0041 &0.8560 $\pm$ 0.0124 &0.8079 $\pm$ 0.0047 &0.8710 $\pm$ 0.0030 &0.8363 $\pm$ 0.0068 &0.7510 $\pm$ 0.0071 &\textbf{0.8991 $\pm$ 0.0012}\\

 & &\cmark &0.8732 $\pm$ 0.0037 &0.7901 $\pm$ 0.0047 &0.8650 $\pm$ 0.0032 &0.9091 $\pm$ 0.0014 &0.8512 $\pm$ 0.0062 &0.8347 $\pm$ 0.0039 &\textbf{0.9316 $\pm$ 0.0015}\\

\cmidrule{2-10}

 &\multirow{2}{*}{\textbf{ICEWS1819}} &\xmark &0.9115 $\pm$ 0.0081 &0.9390  $\pm$ 0.0054 &0.9151 $\pm$ 0.0061 &0.9330  $\pm$ 0.0076 &0.9471 $\pm$ 0.0011 &0.8858 $\pm$ 0.0089 &\textbf{0.9613 $\pm$ 0.0010}\\

 & &\cmark &0.9285 $\pm$ 0.0065 &0.9030 $\pm$ 0.0097 &0.9706 $\pm$ 0.0054 &0.9774 $\pm$ 0.0039 &\textbf{0.9778 $\pm$ 0.0012}  &0.9605  $\pm$ 0.0025 &0.9630  $\pm$ 0.0027\\

 \cmidrule{2-10}

\textit{in.} &\multirow{2}{*}{\textbf{Googlemap CT}} &\xmark &OOM &OOM &0.7958 $\pm$ 0.0012 &\textbf{0.7968 $\pm$ 0.0007} &0.7104 $\pm$ 0.0015 &0.6675 $\pm$ 0.0033 &0.7148  $\pm$ 0.0024\\

 & &\cmark &OOM &OOM &\textbf{0.8791 $\pm$ 0.0028} &0.7058 $\pm$ 0.0047 &0.7895 $\pm$ 0.0046 &0.7543 $\pm$ 0.0018 & 0.7648 $\pm$ 0.0052\\

\cmidrule{2-10}

&\multirow{2}{*}{\textbf{GDELT}} &\xmark &0.8977 $\pm$ 0.0035 &0.8791 $\pm$ 0.0002 &0.7501 $\pm$ 0.0074 &0.7909  $\pm$ 0.0010  & 0.8544$\pm$ 0.0045 &0.7361 $\pm$ 0.0058 &\textbf{0.9135$\pm$ 0.0024}\\

& &\cmark &0.8921 $\pm$ 0.0065 &0.8917 $\pm$ 0.0007 &0.9012 $\pm$ 0.0011  &0.8899 $\pm$ 0.0082 &0.9099 $\pm$ 0.0022 &0.8942 $\pm$ 0.0035  &\textbf{0.9206 $\pm$ 0.0003}\\

\bottomrule
\end{tabular}}
\label{tab:link_prediction}
\end{table*}

\begin{table*}[t]
\caption{Hits@10 for node retrieval. \textit{tr.} means transductive
setting and \textit{in.} means inductive setting. \textbf{Text}
means whether to use Bert-encoded embeddings for initialization.}
\centering \resizebox{1\textwidth}{!}{
\begin{tabular}{c|c|c|ccccccc}
\toprule
\textbf{}
&\textbf{Datasets}
&\textbf{Text}
&\multicolumn{1}{c}{\textbf{JODIE}}
&\multicolumn{1}{c}{\textbf{DyRep}}
&\multicolumn{1}{c}{\textbf{TGAT}}
&\multicolumn{1}{c}{\textbf{CAWN}}
&\multicolumn{1}{c}{\textbf{TCL}}
&\multicolumn{1}{c}{\textbf{GraphMixer}}
&\multicolumn{1}{c}{\textbf{DyGFormer}}\\
\midrule
&\multirow{2}{*}{\textbf{GDELT}} &\xmark &\textbf{0.8733 $\pm$ 0.0095}  &0.8427 $\pm$ 0.0031  &0.7780 $\pm$ 0.0047  &0.7057  $\pm$ 0.0086 &0.8134 $\pm$ 0.0079  &0.7798 $\pm$ 0.0045 &0.8468 $\pm$ 0.0021\\

& &\cmark &0.8675 $\pm$ 0.0101 &0.8399 $\pm$ 0.0037 &0.8817 $\pm$ 0.0035 &0.8747 $\pm$ 0.0074 &0.8875 $\pm$ 0.0030 &0.8678 $\pm$ 0.0076 &\textbf{0.9016 $\pm$ 0.0011}\\

\cmidrule{2-10}

 &\multirow{2}{*}{\textbf{Googlemap CT}} &\xmark &OOM &OOM &\textbf{0.6539 $\pm$ 0.0047} &0.5158 $\pm$ 0.0080 &0.4360 $\pm$ 0.0063 &0.4076 $\pm$ 0.0017 &0.4453  $\pm$ 0.0009\\

 & &\cmark &OOM &OOM &\textbf{0.6972 $\pm$ 0.0022} &0.5239 $\pm$ 0.0063 &0.5373 $\pm$ 0.0057 &0.4855 $\pm$ 0.0028 &0.4913$\pm$ 0.0023\\

 \cmidrule{2-10}

\textit{tr.} &\multirow{2}{*}{\textbf{Amazon movies}} &\xmark   &OOM &OOM &0.6430 $\pm$ 0.0096 &0.5835 $\pm$ 0.0071   &0.6446 $\pm$ 0.0062   &0.6478 $\pm$ 0.0035   &\textbf{0.5055 $\pm$ 0.0084}\\

 & &\cmark  &OOM &OOM   &0.7245 $\pm$ 0.0138   &0.6757 $\pm$ 0.0084   &0.7174 $\pm$ 0.0024   &0.6805 $\pm$ 0.0026   &\textbf{0.7221 $\pm$ 0.0012}\\

\cmidrule{2-10}

&\multirow{2}{*}{\textbf{Yelp}} &\xmark &OOM &OOM   &\textbf{0.5961 $\pm$ 0.0085}  &0.5410 $\pm$ 0.0105   &0.5930 $\pm$ 0.0254   &0.5745 $\pm$ 0.0248   &0.4944 $\pm$ 0.0068\\

 & &\cmark &OOM &OOM &\textbf{0.8069 $\pm$ 0.0136} &0.7648 $\pm$ 0.0088 &0.5745 $\pm$ 0.0197 &0.6585 $\pm$ 0.0064   &0.7600 $\pm$ 0.0036\\

 \midrule

 &\multirow{2}{*}{\textbf{GDELT}} &\xmark &\textbf{0.7393 $\pm$ 0.0079} &0.7285 $\pm$ 0.0027 &0.6453 $\pm$ 0.0085 &0.5752 $\pm$ 0.0038 &0.6157 $\pm$ 0.0097 &0.6200 $\pm$ 0.0065 &0.7086 $\pm$ 0.0056\\

 & &\cmark &0.7344 $\pm$ 0.0064 &0.7266 $\pm$ 0.0061 &0.7329 $\pm$ 0.0029 &0.6925$\pm$ 0.0019 &0.7527 $\pm$ 0.0011 &0.7123 $\pm$ 0.0046 & \textbf{0.7844 $\pm$ 0.0018}\\

\cmidrule{2-10}

 &\multirow{2}{*}{\textbf{Googlemap CT}} &\xmark &OOM &OOM &0.0993 $\pm$ 0.0025    &0.0930 $\pm$ 0.0018   &0.0997 $\pm$ 0.0021   &\textbf{0.1200 $\pm$ 0.0002}  &0.0620 $\pm$ 0.0019\\

 & &\cmark &OOM &OOM &\textbf{0.3705 $\pm$ 0.0023} &0.2236 $\pm$ 0.0076 &0.1769 $\pm$ 0.0009 &0.1307 $\pm$ 0.0011 &0.1397 $\pm$ 0.0012\\

 \cmidrule{2-10}

\textit{in.} &\multirow{2}{*}{\textbf{Amazon movies}} &\xmark   &OOM &OOM  &0.5559 $\pm$ 0.0125    &0.4716 $\pm$ 0.0102   &\textbf{0.5559 $\pm$ 0.0051}  &0.5548 $\pm$ 0.0039   &0.4617 $\pm$ 0.0138\\

 & &\cmark &OOM &OOM &\textbf{0.6461 $\pm$ 0.0093} &0.5818 $\pm$ 0.0073 &0.6382 $\pm$ 0.0031 &0.5915 $\pm$ 0.0024 &0.6481 $\pm$ 0.0016 \\

\cmidrule{2-10}

&\multirow{2}{*}{\textbf{Yelp}} &\xmark &OOM &OOM &\textbf{0.5282 $\pm$ 0.0160}    &0.4624 $\pm$  0.0138  &0.5245 $\pm$ 0.0170   &0.4832 $\pm$ 0.0297   &0.4265 $\pm$ 0.0103\\

& &\cmark &OOM &OOM &\textbf{0.7204 $\pm$ 0.0130}  &0.6777 $\pm$ 0.0104   &0.4832 $\pm$ 0.0134   &0.5718 $\pm$ 0.0084   &0.6825 $\pm$ 0.0036\\

\bottomrule
\end{tabular}}
\label{tab:node_retrieval}
\end{table*}

\begin{figure}[t]
\centering
\subfigure[GDELT]{\includegraphics[width=0.48\linewidth]{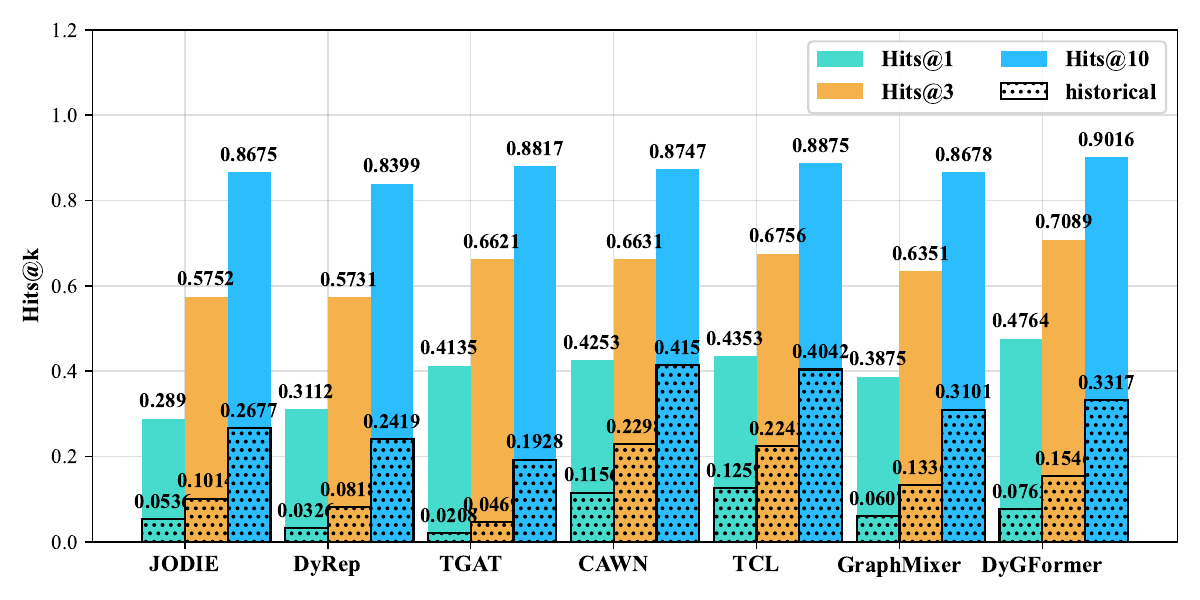}}
\subfigure[Googlemap
CT]{\includegraphics[width=0.48\linewidth]{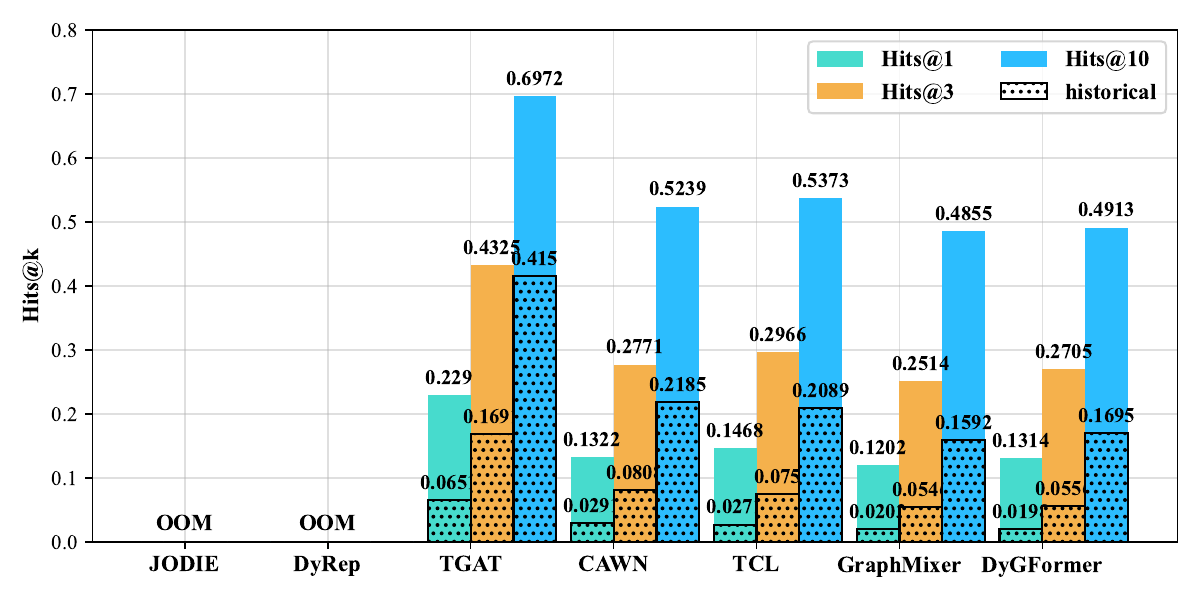}}
\caption{Node retrieval performance using random sampling and
historical sampling.} \label{fig:node_retrieval_sample_ablation}
\end{figure}

\begin{table}[!ht]
\centering \caption{Precision, Recall and F1 of BERTscore of
different LLMs for textural relation generation. The number of test
samples is 500 per dataset.} \label{tab:text_generation}
\resizebox{0.75\linewidth}{!}{
\begin{tabular}{l|ccc|ccc|ccc}\toprule
 & \multicolumn{3}{c|}{\textbf{Googlemap CT}} & \multicolumn{3}{c|}{\textbf{Amazon movies}} & \multicolumn{3}{c}{\textbf{Stack elec}} \\
 & \multicolumn{1}{c}{Precision} & \multicolumn{1}{c}{Recall} & \multicolumn{1}{c|}{F1} & \multicolumn{1}{c}{Precision} & \multicolumn{1}{c}{Recall} & \multicolumn{1}{c|}{F1} & \multicolumn{1}{c}{Precision} & \multicolumn{1}{c}{Recall} & \multicolumn{1}{c}{F1} \\ \midrule
GPT 3.5 turbo & 79.89 & \textbf{84.13} & 81.91 & 79.79 & 83.61 & 81.63 & 80.52 & 81.96 & 81.21 \\
GPT 4o & 78.33 & 84.06 & 81.07 & 78.68 & \textbf{84.20} & 81.33 & 78.30 & 82.37 & 80.26 \\
Llama3-8b & 78.62 & 83.84 & 81.12 & 78.48 & 83.97 & 81.09 & 79.91 & 82.35 & 81.09 \\
Mistral-7b & \textbf{80.21} & 84.05 & \textbf{82.07} & 79.81 & 84.05 & \textbf{81.84} & 80.25 & \textbf{82.61} & 81.40 \\
Vicuna-7b & 80.04 & 83.79 & 81.85 & \textbf{80.23} & 83.60 & 81.83 & \textbf{80.65} & 82.37 & \textbf{81.46} \\
Vicuna-13b & 80.14 & 84.00 & 81.99 & 77.59 & 83.56 & 80.39 & 80.57 & 82.20 & 81.33 \\
\bottomrule
\end{tabular}}
\end{table}

\subsection{Destination Node Retrieval}
\label{sec:node_retrieval}

Table~\ref{tab:node_retrieval} shows the performance of the node
retrieval task. For each test example, 1000 nodes including the
ground truth node are randomly sampled to create the candidate set
for ranking. We observe that existing models fail to achieve
satisfactory performance, showing their weakness in accurately
capturing the dynamic interaction preferences of nodes. Text
attributes improve the model performance both in the transductive
and inductive settings. This is because the descriptions and
historical interactions with text can reflect the preference of a
node (\eg reviews from a user reflect personalized opinion and
preference). To further investigate the performance of existing
models, we sample 1000 nodes that are the most recently interacted
as the candidate set for each test node, which presents a more
challenging setting (denoted as \textit{historical sampling}). As
shown in Figure~\ref{fig:node_retrieval_sample_ablation}, existing
models perform significantly worse in the historical sampling
setting, since these models largely rely on capturing structural and
temporal co-occurrences, but ignore the semantic relevance and
long-range dependencies that are important in real-world
applications. More experimental results are provided in
Appendix~\ref{appd:node_retrieval_result}.

\subsection{Textural Relation Generation}

\begin{wraptable}{r}{0.45\textwidth}
\vspace{-4mm} \centering \caption{Performance of LLMs after SFT for
the relation generation task in terms of BERTscore (F1).}
\label{tab:finetune} \resizebox{0.9\linewidth}{!}{
\begin{tabular}{lll}\toprule
 & \textbf{Googlemap CT} & \textbf{Stack elec} \\ \midrule
Llama3-8b & 81.12 & {81.09} \\
Llama3-8b + SFT & 81.84 (0.72$\uparrow$) & 81.97 (0.88$\uparrow$)\\
\midrule
Vicuna-7b & 81.85 & {81.46} \\
Vicuna-7b + SFT & \textbf{85.67} (3.82$\uparrow$)& 82.67
(1.21$\uparrow$)\\ \midrule
Vicuna-13b & 81.99 & {81.33} \\
Vicuna-13b + SFT & 84.67 (2.68$\uparrow$)& \textbf{82.73}
(1.40$\uparrow$) \\ \bottomrule
\end{tabular}}
\end{wraptable}

Generating the textual content of future interactions within certain
node pairs remains an under-explored challenge, which requires a
language model to understand the dynamics and textural description
within a graph structure. We evaluate six LLMs on three datasets
derived from different real-world scenarios. For instance, on the
\texttt{Googlemap CT} dataset, the input sentence is constructed by
the user's historical reviews and textural description of the
destination. Then, LLM is prompted to generate potential reviews
from the user to the future destination. Instead, the LLM is
prompted to generate reviews for target movies and answers to target
questions on the \texttt{Amazon movies} and \texttt{Stack elec}
datasets, respectively. See Appendix~\ref{appd:prompt} for
dataset-specific prompts. As shown in
Table~\ref{tab:text_generation}, we observe that open-source LLMs
such as Mistral and Vicuna perform comparably well to proprietary
LLMs in this task. To further improve their performance on relation
generation, we extract 10,000 node pair interactions associated with
textural descriptions from the datasets for LLM supervised
fine-tuning (SFT). Results in Table~\ref{tab:finetune} demonstrate
that LLMs consistently achieve enhanced performance in textural
generation after supervised fine-tuning. Especially, Vicuna-7b
benefits the most from SFT. We provide the ablation study on
different information provided to LLM in
Appendix~\ref{appd:relation_generation_result}.

\section{Discussion}

\xhdr{Limitations and Future Directions} While the proposed DTGB
represents a significant advancement in the study of DyTAGs, there
are areas ripe for further exploration. Our extensive benchmark
experiments reveal that current dynamic graph learning algorithms
and large language models (LLMs) exhibit varying degrees of
effectiveness when handling the complex interactions between the
dynamic graph structure and textural attributes. This finding
highlights the potential for even greater improvements and
innovations in this field in the future.

A particularly exciting future direction is the design of temporal
graph tokens that can directly incorporate dynamic graph information
into LLMs for reasoning and dynamics-aware generation. By designing
representations that seamlessly blend structural and temporal
aspects of the graphs with their text attributes, these tokens could
potentially enhance the ability of LLMs to capture and utilize the
dynamic nature of DyTAGs. This approach promises to improve
performance in a range of applications, such as real-time
recommendation systems, dynamic knowledge graphs, and evolving
social network analysis.

Another notable challenge is the scalability issue when handling
large-scale DyTAGs, especially given the potentially long text
descriptions associated with nodes and edges. The complexity of
encoding long sequences and integrating them with dynamic graph
structures can lead to computational overhead. Addressing this
scalability issue is a crucial future direction to ensure that
models can efficiently process large-scale graphs with extensive
textual attributes, paving the way for more practical and robust
applications in real-world scenarios.

\xhdr{Broader Impact} The broader impact of DTGB lies in its ability
to drive advancements in dynamic text-attributed graph research by
providing a comprehensive benchmark for evaluating models. The
broader impact can extend to numerous societal and technological
domains, such as social media and real-time recommendation systems.
Furthermore, advancements driven by DTGB that can integrate dynamic
graph learning with natural language processing could lead to
methodological enhancement in fields such as healthcare, finance,
and cybersecurity, where understanding the evolving relationships
and information is critical for decision-making and risk management.
Overall, DTGB has the potential to drive significant improvements in
how complex, dynamic data is harnessed and utilized across various
sectors.

\section{Conclusion}
\label{sec:conclusion}

We propose the first comprehensive benchmark DTGB specifically for
dynamic text-attributed graphs (DyTAGs). We collect and provide
eight carefully processed dynamic text-attributed graph datasets
from diverse domains. Based on these datasets, we comprehensively
investigate the performance of existing dynamic graph learning
models and large language models (LLMs) in four real-world-driven
tasks. Our experimental results validate the utility of DTGB and
provide insights for further technical advancements. The limitation
of this work is that we did not incorporate the high-order graph
context in the textual relation generation task, due to the maximum
input length of LLMs. Therefore, in the future, we will investigate
how to efficiently use LLM to handle high-order dynamic topology and
long-range evolving texts within DyTAGs.

\begin{ack}
This work is supported by the Shenzhen Science and Technology
Program (No. JCYJ20210324121213037) and Guangxi Key Research and
Development Program (No. Guike AB24010112).
\end{ack}

{\small
\bibliographystyle{unsrt}
\bibliography{reference}
}
\clearpage
\section*{Checklist}
%%% BEGIN INSTRUCTIONS %%%
%The checklist follows the references.  Please
%read the checklist guidelines carefully for information on how to answer these
%questions.  For each question, change the default \answerTODO{} to \answerYes{},
%\answerNo{}, or \answerNA{}.  You are strongly encouraged to include a {\bf
%justification to your answer}, either by referencing the appropriate section of
%your paper or providing a brief inline description.  For example:
%\begin{itemize}
%  \item Did you include the license to the code and datasets? \answerYes{See Section~\ref{gen_inst}.}
%  \item Did you include the license to the code and datasets? \answerNo{The code and the data are proprietary.}
%  \item Did you include the license to the code and datasets? \answerNA{}
%\end{itemize}
%Please do not modify the questions and only use the provided macros for your
%answers.  Note that the Checklist section does not count towards the page
%limit.  In your paper, please delete this instructions block and only keep the
%Checklist section heading above along with the questions/answers below.
%%% END INSTRUCTIONS %%%

\begin{enumerate}

\item For all authors...
\begin{enumerate}
  \item Do the main claims made in the abstract and introduction accurately reflect the paper's contributions and scope?
    \answerYes{As we have mentioned in the contribution summarization of Section~\ref{sec:introduction}, we propose the first open benchmark specifically designed for dynamic text-attributed graphs, and provide a comprehensive evaluation of existing models.}
  \item Did you describe the limitations of your work?
    \answerYes{As we have mentioned in Section~\ref{sec:conclusion}, the limitation of this work is that we did not incorporate the high-order graph context in the textual relation generation task, due to the maximum input length of LLMs.}
  \item Did you discuss any potential negative societal impacts of your work?
    \answerNA{This work has no potential negative societal impacts.}
  \item Have you read the ethics review guidelines and ensured that your paper conforms to them?
    \answerYes{}
\end{enumerate}

\item If you are including theoretical results...
\begin{enumerate}
  \item Did you state the full set of assumptions of all theoretical results?
    \answerNA{This work does not include theoretical results.}
    \item Did you include complete proofs of all theoretical results?
    \answerNA{This work does not include theoretical results.}
\end{enumerate}

\item If you ran experiments (e.g. for benchmarks)...
\begin{enumerate}
  \item Did you include the code, data, and instructions needed to reproduce the main experimental results (either in the supplemental material or as a URL)?
    \answerYes{We have provided the detailed implementation of different models in Appendix~\ref{appd:baseline}. The dataset and source code are available at \url{https://github.com/zjs123/DTGB}.}
  \item Did you specify all the training details (e.g., data splits, hyperparameters, how they were chosen)?
    \answerYes{We have provided the detailed implementation and hyperparameter settings of different models in Appendix~\ref{appd:baseline}.}
    \item Did you report error bars (e.g., with respect to the random seed after running experiments multiple times)?
    \answerYes{They are illustrated in the tables in Section~\ref{sec:experiment}.}
    \item Did you include the total amount of compute and the type of resources used (e.g., type of GPUs, internal cluster, or cloud provider)?
    \answerYes{}
\end{enumerate}

\item If you are using existing assets (e.g., code, data, models) or curating/releasing new assets...
\begin{enumerate}
  \item If your work uses existing assets, did you cite the creators?
    \answerYes{}{}
  \item Did you mention the license of the assets?
    \answerYes{}
  \item Did you include any new assets either in the supplemental material or as a URL?
    \answerYes{}
  \item Did you discuss whether and how consent was obtained from people whose data you're using/curating?
    \answerNA{All of our used raw data are publicly available.}
  \item Did you discuss whether the data you are using/curating contains personally identifiable information or offensive content?
    \answerNA{Our raw data contains no personally identifiable information or offensive content.}
\end{enumerate}

\item If you used crowdsourcing or conducted research with human subjects...
\begin{enumerate}
  \item Did you include the full text of instructions given to participants and screenshots, if applicable?
    \answerNA{We did not use crowdsourcing.}
  \item Did you describe any potential participant risks, with links to Institutional Review Board (IRB) approvals, if applicable?
    \answerNA{We did not use crowdsourcing.}
  \item Did you include the estimated hourly wage paid to participants and the total amount spent on participant compensation?
    \answerNA{We did not use crowdsourcing.}
\end{enumerate}

\end{enumerate}

\clearpage
\appendix
\section{Datasets}

\subsection{Dataset Description}
\label{appd:dataset_intro}

\xhdr{Dataset Format} For each dataset in DTGB, we provide three
different files. We store each edge as a tuple in the edge\_list.csv
file, which includes id of the source node, id of the target node,
id of the relation between them, the occurring timestamp, and the
edge category. We use entity\_text.csv and relation\_text.csv to
store the text attributes of nodes and edges for each dataset. Each
file includes the mapping from node and relation ids to the
corresponding raw text descriptions. All the datasets and codes to
reproduce the results in this paper are available at
\url{https://github.com/zjs123/DTGB}. We provide a detailed
description and the link to the raw resources of each dataset as
follows.

\xhdr{Enron\footnote{\url{https://www.cs.cmu.edu/~enron/}}} This
dataset is derived from the email communications between employees
of the ENRON energy corporation over three years (1999-2002). The
nodes indicate the employees while the edges are e-mails among them.
The text attribute of each node is extracted from the department and
position of the employee (if available). The text attribute of each
edge is the raw text of e-mails. Non-English statements, abnormal
symbols, and tables are removed from the raw text and we perform
length truncation on these e-mails. The edge categories are
extracted from the e-mail archive of the raw resource. There are 10
kinds of categories such as \textit{calendar}, \textit{notes}, and
\textit{deal communication}. We order edges in this dataset based on
the sending timestamps of e-mails.

\xhdr{GDELT\footnote{\url{https://www.gdeltproject.org/}}} This
dataset is derived from the Global Database of Events, Language, and
Tone project, which is an initiative to construct a catalog of
political behavior across all countries of the world. Nodes in this
dataset indicate political entities such as \textit{United States}
and \textit{Kim Jong UN}. We directly use the names of these
entities as their textual attributes. Edges in this dataset
represent the relationships between entities (e.g.,
\textit{Prisident of} and \textit{Make Statement}). We use the
descriptions of these relationships as the textual attributes of
edges. Each edge category refers to a kind of political relationship
or behavior. We order edges in this dataset based on the occurring
timestamps of these political events.

\xhdr{ICEWS1819\footnote{\url{https://dataverse.harvard.edu/dataverse/icews}}}
This dataset is derived from the Integrated Crisis Early Warning
System project, which is also a temporal knowledge graph for
political events. We extract events from 2018-01-01 to 2019-12-31 to
construct this dataset. We organize the name, sector, and
nationality of each political entity as its text attribute, while
the edge text attributes are the descriptions of the political
relationships. The edge categories refer to the types of political
relationships or behavior. Similar to the GDELT dataset, we order
edges in this dataset based on the occurring timestamps of the
political events. Note that the major difference between the
ICEWS1819 and GDELT datasets is that first, the time granularity of
GDELT is 15 minutes, while that of ICEWS1819 is 24 hours. Therefore
GDELT describes political interactions in a more fine-grained way.
Second, ICEWS1819 has a 4 times larger node set compared with GDELT
and thus represents a more sparse scenario.

\xhdr{Stack
elec\footnote{\url{https://archive.org/details/stackexchange}}}
Stack Exchange Data is an anonymized dump of all user-contributed
content on various stack exchange sites. It includes questions,
answers, comments, tags, and other related data from these sites. We
regard the questions and users in these sites as nodes, while the
answers and comments from users to questions are regarded as
text-attributed edges, subsequently constructing a dynamic bipartite
graph that describes the multi-round dialogue between users and
questions. We extract all the questions related to electronic
techniques as well as the corresponding answers and comments to
construct the Stack elec dataset. For user nodes, we use the
self-introductions of users as their text attributes, which describe
the technical areas that the user is familiar with. For the question
node, we use the title and the body of each question post as its
text attribute. We use the raw text of answers and comments as the
text attributes of edges. We construct two categories based on the
voting of each answer: \textit{Useful} if the voting count is larger
than 1, otherwise \textit{Useless}. We order edges in this dataset
based on the answering timestamps from users.

\xhdr{Stack
ubuntu\footnote{\url{https://archive.org/details/stackexchange}}}
This is another dataset from Stack Exchange Data, which contains all
the questions related to the Ubuntu system. Besides the size and the
topic, the biggest difference between this dataset and \texttt{Stack
elec} is that the answers in this dataset are usually a mixture of
codes and natural language, which brings more challenges to the
understanding of the semantic context of interactions.

\xhdr{Googlemap
CT\footnote{\url{https://datarepo.eng.ucsd.edu/mcauley_group/gdrive/googlelocal/}}}
This dataset is extracted from the Google Local Data project, which
contains review information on Google map as well as the user and
business information up to September 2021 in the United States. We
extract all the business entities from Connecticut State to
construct this dataset. Nodes are users and business entities while
edges are reviews from users to businesses. Only the business nodes
are enriched with text attributes, containing the name, address,
category, and self-introduction of the business entity. The edge
text attributes are the raw text of user reviews. The edge
categories are integers from 1 to 5, derived from the ratings from
users to businesses. We have removed emojis and meaningless
characters from reviews. Edges in this dataset are ordered based on
the review timestamps from users.

\xhdr{Amazon
movies\footnote{\url{https://cseweb.ucsd.edu/~jmcauley/datasets/amazon_v2/}}}
This dataset is extracted from the Amazon Review Data project, which
contains product reviews and metadata from Amazon spanning May 1996
to July 2014. To construct this dataset, we extract products in the
class of \textit{Movies and TV} and the corresponding reviews. The
text attribute of each product node contains its name, category,
description, and rank score. The text attributes of edges are review
text from users to products. Similarly, the edge categories are
integers from 1 to 5, derived from the ratings from users to
businesses. We still order edges in this dataset based on the review
timestamps from users.

\xhdr{Yelp\footnote{\url{https://www.yelp.com/dataset}}} This
dataset is extracted from the Yelp Open Dataset project which
contains reviews of restaurants, shopping centers, hotels, tourism,
and other businesses from users. The text attribute of each business
node contains its name, address, city, and category. The text
attribute of each user node contains its first name, number of
reviews, and register time. Edge text is the reviews from users to
businesses. Edge categories are also the ratings from 1 to 5. All
the edges are ordered based on the review timestamps from users.

\subsection{Dataset Analysis}
\label{appd:dataset_analysis}

\begin{figure}[t]
  \centering
  \includegraphics[width=1.0\linewidth]{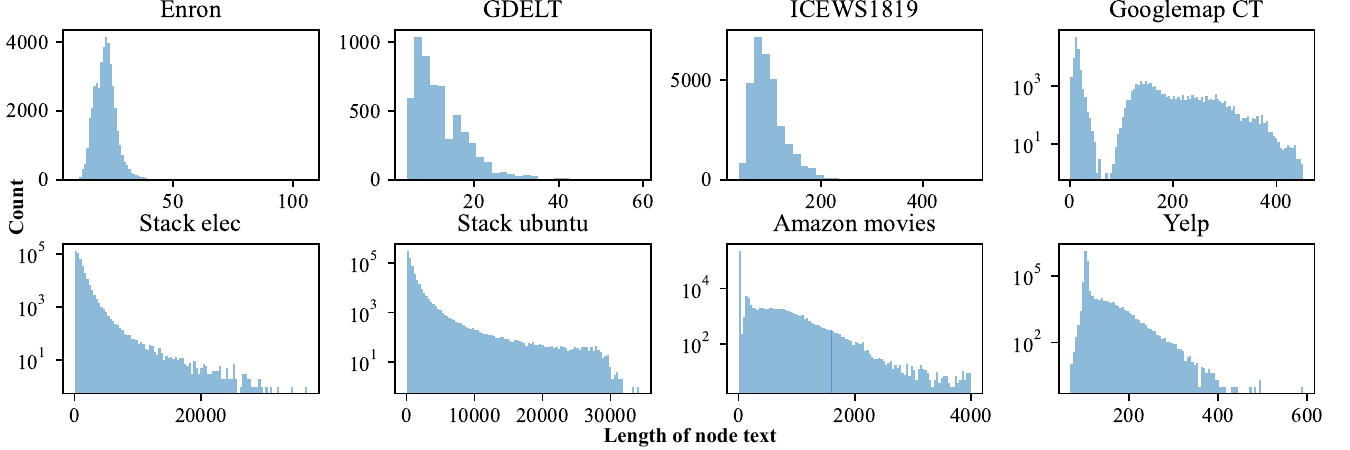}
  \caption{Distribution of node text length on DTGB datasets.}
  \label{fig:text_length_node}
\end{figure}
\begin{figure}[t]
  \centering
  \includegraphics[width=1.0\linewidth]{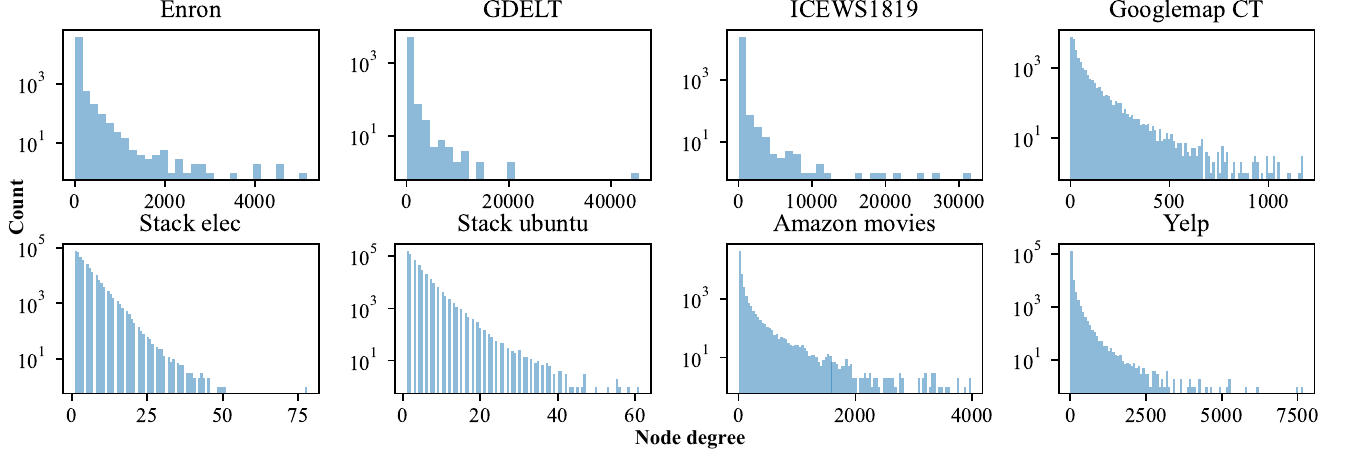}
  \caption{Distribution of node degree on DTGB datasets.}
  \label{fig:node_degree}
\end{figure}
\begin{figure}[t]
  \centering
  \includegraphics[width=1.0\linewidth]{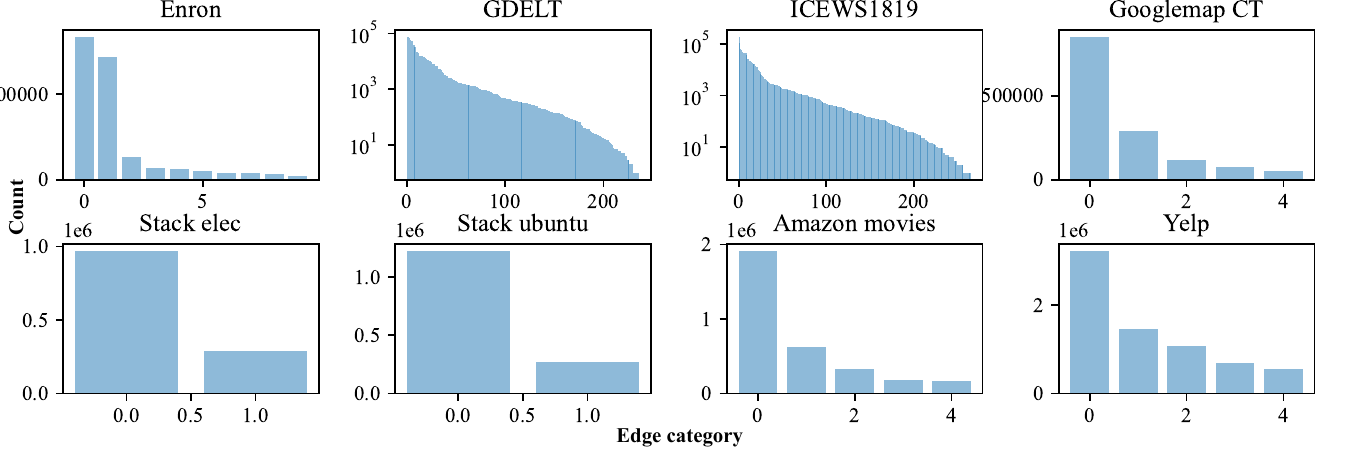}
  \caption{Distribution of the number of edges for each category on DTGB datasets.}
  \label{fig:cat_num}
\end{figure}

Here, we provide more statistical analysis for datasets in the DTGB
benchmark. As illustrated in Figure~\ref{fig:text_length_node}, we
can see that datasets \texttt{Stack elec} and \texttt{Stack ubuntu}
have significantly longer node text than other datasets. This is
because question text in these datasets usually contains references
and the introduction of background, which will bring unique
challenges in understanding the node semantics. One interesting
observation is that the distribution of the \texttt{Googlemap CT}
dataset is bimodal. This is because some businesses in this dataset
lack \textit{description} raw text, and thus have significantly
shorter text length than others. In Figure~\ref{fig:node_degree}, we
can see that most datasets in the DTGB benchmark show long-tail
distribution on node degrees, which meets the real world. In
Figure~\ref{fig:cat_num}, we can see that the categories in these
datasets are non-uniformly distributed. We preserve such skewed
distribution to faithfully reflect the challenges in real-world
applications and leave opportunities to handle these challenges
through the incorporation of dynamic graph structure and text
semantic modeling.

\subsection{License}

All the used codes and datasets are publicly available and permit
usage for research purposes under either MIT License or Apache
License 2.0.

\section{Notations}
\label{appd:notion}

\begin{table}[t] \small
 \centering
  \caption{Important notations and descriptions.}
  \label{tab:notation}
  \begin{tabular}{cc}
    \toprule
    Notation & Description\\
    \midrule
    $\mathcal{G}$ & A dynamic text-attributed graph.\\
    $\mathcal{V}, \mathcal{E}$ & Node and edge sets of $\mathcal{G}$. \\
    $\mathcal{G}_{T}$ & Subgraph of $\mathcal{G}$ which contains nodes and edges appeared before timestamp $T$.\\
    $\mathcal{V}_{T}, \mathcal{E}_{T}$ & Node and edge sets of $\mathcal{G}_{T}$.\\
    \hline
    $\mathcal{D}$ & Set of node text descriptions. \\
    $\mathcal{R}$ & Set of edge text descriptions. \\
    $\mathcal{L}$ & Set of edge categories. \\
    $\mathcal{T}$ & Set of observed timestamps. \\
    \hline
    $u, v$ & Nodes in the dynamic text-attributed graph. \\
    $(u, v)$ & An edge in the dynamic text-attributed graph which connects nodes $u$ and $v$. \\
    $d_v$ & Text description of node $v$. \\
    $r_{u, v}$ & Text description of edge $(u, v)$. \\
    $l_{u, v}$ & Category of edge $(u, v)$. \\
    $t_{u, v}$ & Occurring timestamp of edge $(u, v)$. \\
    \bottomrule
  \end{tabular}
\end{table}

In Table~\ref{tab:notation}, we summarize important notations used
in this paper and provide the corresponding descriptions.

\section{Experiments}

\subsection{Baselines}
\label{appd:baseline}

\subsubsection{Temporal Graph Models}
\label{appd:TGB_model}

\xhdr{JODIE \cite{JODIE}} This model learns to project/forecast the
embedding trajectories into the future to make predictions about the
entities and their interactions. Two coupled recurrent neural
networks are used to update the states of entities and a projection
operation is used to get the future representation trajectory of
each entity.

\xhdr{DyRep \cite{dyrep}} This model proposes a recurrent
architecture to update node states upon each interaction. It uses a
deep temporal point process model to capture the dynamics of the
observed processes. This model is further parameterized by a
temporal-attentive network that encodes temporally evolving
structural information into node representations which in turn
drives the nonlinear evolution of the observed graph dynamics.

\xhdr{TGAT \cite{TGAT}} This model aims to efficiently aggregate
temporal-topological neighborhood features as well as to learn the
time-feature interactions. It uses the self-attention mechanism as a
building block. A functional time encoding technique based on the
classical Bochner's theorem from harmonic analysis is used to
capture temporal patterns.

\xhdr{CAWN \cite{CAWN}} This model adopts an anonymization strategy
based on a set of sampled walks to explore the causality of network
dynamics and generate inductive node identities. Then, a neural
network model is used to encode these sampled walks and aggregate
them to obtain the final node representation.

\xhdr{TCL \cite{TCL}} This model designs a two-stream encoder that
separately processes temporal neighborhoods associated with the two
target interaction nodes. Then, a graph-topology-aware Transformer
is proposed to consider both graph topology and temporal information
to learn node representations. Cross-attention operation is also
incorporated to learn the relevance between two interaction nodes.

\xhdr{GraphMixer \cite{Graphmixer}} This model shows the
effectiveness of the fixed-time encoding function in modeling
dynamic interactions. The proposed simple architecture consists of
three components: a link encoder used to summarize the information
from temporal links, a node encoder used to summarize node
information, and a link classifier that performs link prediction.

\xhdr{DyGFormer \cite{DyGLib}} This model learns node
representations from nodes' historical first-hop interactions. It
uses a neighbor co-occurrence encoding scheme to explore the
correlations between nodes based on their historical sequences. A
patching technique is also proposed to divide each sequence into
multiple patches, which allows the model to effectively benefit from
longer histories.

\subsubsection{Large Language Models}
\label{appd:LLM}

\xhdr{Mistral (7B) \cite{jiang2023mistral}} This language model
leverages grouped-query attention (GQA) for faster inference,
coupled with sliding window attention (SWA) to effectively handle
sequences of arbitrary length with a reduced inference cost. We use
the Mistral-7B-Instruct-v0.2 version in our experiment.

\xhdr{Llama-3 (8B) \cite{llama3}} This is an auto-regressive
language model that uses an optimized Transformer architecture. The
tuned versions use supervised fine-tuning (SFT) and reinforcement
learning with human feedback (RLHF) to align with human preferences
for helpfulness and safety. We use the 8-billion-parameter
instruction version of Llama-3 in our experiment.

\xhdr{Vicuna (7B/13B) \cite{vicuna2023}} Vicuna is fine-tuned from
Llama 2 with supervised instruction fine-tuning. The training data
is from the user-shared conversations collected on the internet. We
use the vicuna-7b-v1.5 version and vicuna-13b-v1.5 version in our
experiment.

\xhdr{GPT3.5-turbo and GPT4o\footnote{\url{https://openai.com/}}}
Generative pre-trained Transformer 3.5 (GPT-3.5) is a sub-class of
GPT-3 models created by OpenAI in 2022. GPT-4o was released on 13
May 2024, which achieves state-of-the-art results in voice,
multilingual, and vision benchmarks.

\subsection{Metrics}
\label{appd:metirc}

In this section, we describe the evaluation metrics used in four
benchmark tasks. For the edge classification task, we use the
weighted average of different categories to avoid the influence of
imbalance category distribution. For future link prediction, we
follow the previous work \cite{DyGLib} to use the Average Precision
and AUC-ROC metrics. For the destination node retrieval task, we
employ the widely used Hits@k metric. For the textual relation
generation task, we use the Bertscore to concentrate on semantic
relevance between ground truth and the generated sentence.

\subsubsection{Edge Classification}
\label{appd:edge_classification}

\xhdr{Weighted Precision} This metric is the weighted average of the
precisions of different classes. The weights are typically the
number of instances for each class or other meaningful metrics.
\begin{equation}
\text{Weighted Precision} = \frac{\sum_{i=1}^{n} w_i \cdot P_i}{\sum_{i=1}^{n} w_i},
\end{equation}
where $w_i$ is the weight of the $i$-th class, and $P_i$ is the
precision of the $i$-th class.

\xhdr{Weighted Recall} This metric is the weighted average of the
recalls of different classes, similar to Weighted Precision.
\begin{equation}
\text{Weighted Recall} = \frac{\sum_{i=1}^{n} w_i \cdot R_i}{\sum_{i=1}^{n} w_i},
\end{equation}
where $w_i$ is the weight of the $i$-th class, and $R_i$ is the
recall of the $i$-th class.

\xhdr{Weighted F1-score} This metric is the weighted average of the
F1-scores of different classes, combining the benefits of Precision
and Recall.
\begin{equation}
\text{Weighted F1-score} = \frac{\sum_{i=1}^{n} w_i \cdot F1_i}{\sum_{i=1}^{n} w_i},
\end{equation}
where $w_i$ is the weight of the $i$-th class, and $F1_i$ is the
F1-score of the $i$-th class, calculated as:
\begin{equation}
F1_i = 2 \cdot \frac{P_i \cdot R_i}{P_i + R_i}.
\end{equation}

\subsubsection{Future Link Prediction}
\label{appd:link_prediction}

\xhdr{Average Precision} This metric is the average of precision
values at different recall levels. It is the area under the
Precision-Recall curve.
\begin{equation}
\text{Average Precision} = \sum_{k=1}^{n} (R_k - R_{k-1}) P_k,
\end{equation}
where $R_k$ is the recall at the $k$-th threshold, and $P_k$ is the
precision at the $k$-th threshold.

\xhdr{AUC-ROC} AUC-ROC (Area Under the Receiver Operating
Characteristic Curve) is a performance measurement for
classification problems at various threshold settings. The ROC is a
probability curve, and AUC represents the degree or measure of
separability.
\begin{equation}
\text{AUC-ROC} = \int_{0}^{1} \text{TPR}(\text {FPR}) \, d(\text{FPR}),
\end{equation}
where $\text{TPR}$ is the true positive rate, and $\text{FPR}$ is
the false positive rate. The ROC curve is created by plotting the
TPR against the FPR at various threshold settings.

\subsubsection{Destination Node Retrieval}
\label{appd:node_retrieval}

\xhdr{Hits@k} Hits@k (Hits at $k$) is a metric used in information
retrieval and recommendation systems to evaluate the effectiveness
of a model in retrieving relevant items. It measures the proportion
of times the true positive item is found within the top-$k$
predictions.
\begin{equation}
\text{Hits@k} =\sum_{i} \frac{\mathbb{I} \left(\operatorname{rank}_{i}\le k\right) }{Q},
\end{equation}
where $\mathbb{I}(*)$ is the indicator function (returns 1 if the
condition is true, otherwise 0), and $Q$ represents the total number
of test samples.

\subsubsection{Textual Relation Generation}
\label{appd:relation_generation}

\xhdr{Bertscore} This metric evaluates the semantic similarity
between candidate and reference texts using the BERT model.
\begin{equation}
\text{BERTscore}(c, r) = \frac{1}{|c|} \sum_{i=1}^{|c|}
\max_{j=1,\ldots,|r|} \text{BERT}(c_i, r_j),
\end{equation}
where $c$ and $r$ are the candidate and reference texts,
respectively, and $\text{BERT}(c_i, r_j)$ is the similarity score
between the $i$-th word in the candidate text and the $j$-th word in
the reference text, computed by the BERT model.

\xhdr{Precision} In the context of BERTscore, precision measures the
accuracy of the generated text by assessing the proportion of
relevant text generated among all the text produced by the model. A
high precision indicates that the generated text is highly relevant
and accurate compared with the reference text. The complete score
matches each token in $\hat{x}$ to a token in $x$ to compute
precision. Then, we use greedy matching to maximize the matching
similarity score, where each token is matched to the most similar
token in the other sentence. The precision score for BERTscore is
calculated as:
\begin{equation}
P_{\mathrm{BERT}}=\frac{1}{|\hat{x}|} \sum_{\hat{x}_j \in \hat{x}}
\max _{x_i \in x} \mathbf{x}_i^{\top} \hat{\mathbf{x}}_j.
\end{equation}

\xhdr{Recall} In the context of BERTscore, recall evaluates the
completeness of the generated text by measuring the proportion of
relevant text captured by the model among all the relevant text
present in the reference text. A high recall indicates that the
model is good at capturing relevant information from the reference
text in its generated output. The complete score matches each token
in $x$ to a token in $\hat{x}$ to compute recall and it is
calculated as:
\begin{equation}
R_{\mathrm{BERT}}=\frac{1}{|x|} \sum_{x_i \in x} \max _{\hat{x}_j
\in \hat{x}} \mathbf{x}_i^{\top} \hat{\mathbf{x}}_j.
\end{equation}

\xhdr{F1-score} In the context of BERTscore, F1-score is the
harmonic mean of precision and recall. It balances between precision
and recall, providing a single metric to assess the overall
performance of the model in generating relevant text. The F1-score
for BERTscore is calculated as:
\begin{equation}
F_{\mathrm{BERT}}=2 \frac{P_{\mathrm{BERT}} \cdot
R_{\mathrm{BERT}}}{P_{\mathrm{BERT}}+R_{\mathrm{BERT}}}.
\end{equation}

\subsection{Prompts for Textual Relation Generation}
\label{appd:prompt}

\begin{figure}[t]
  \centering
  \includegraphics[width=0.85\linewidth]{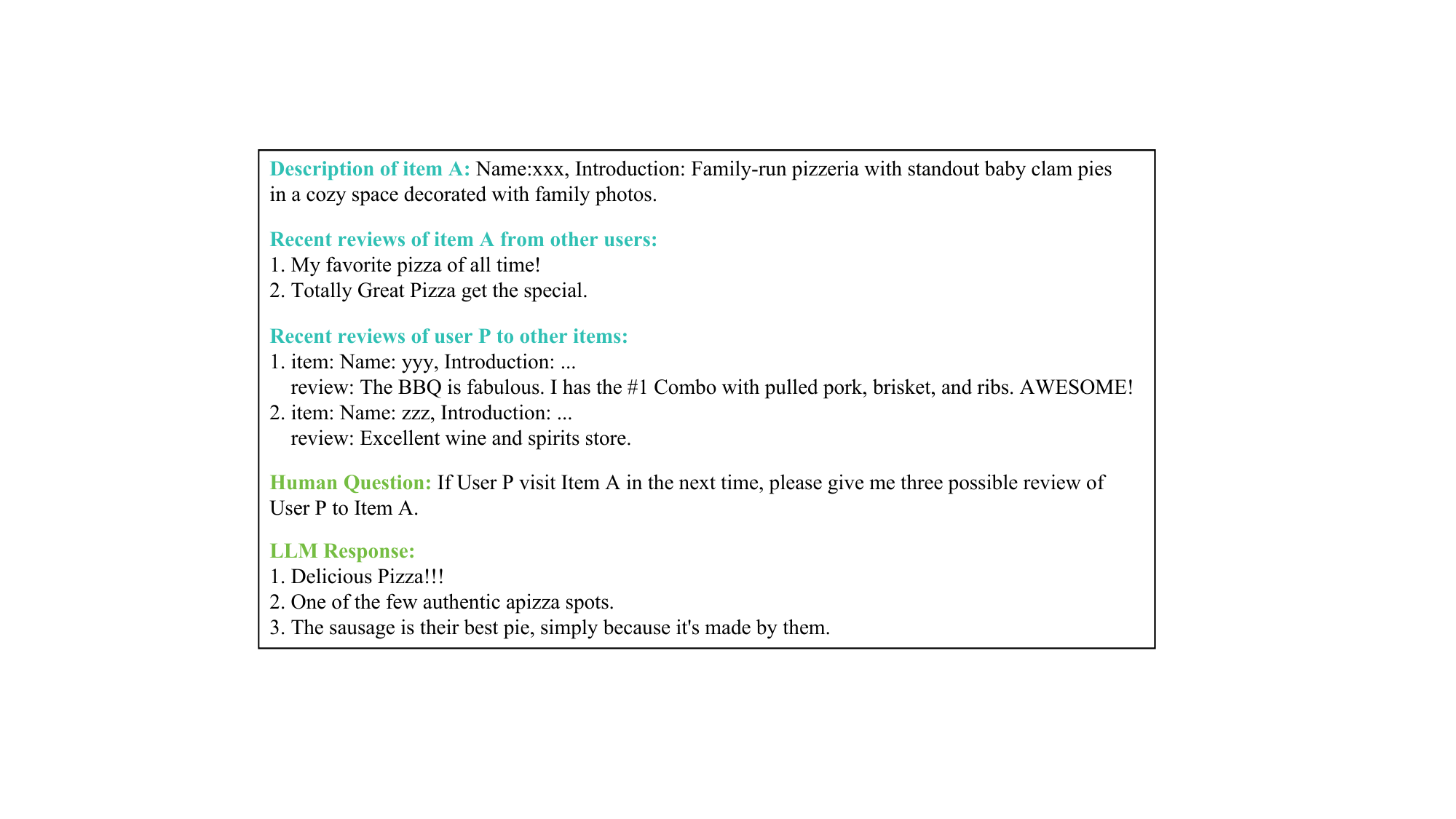}
  \caption{Example of prompts used for inference and fine-tuning in the textual relation generation task (a case in the \texttt{Googlemap CT} dataset).}
  \label{fig:prompt}
\end{figure}

As illustrated in Figure~\ref{fig:prompt}, we give an example of
prompts used in the textual relation generation task, which visually
illustrates the presentation of the instruction to the language
model. Note that for different datasets, some keywords in the prompt
will be consequently changed. For the \texttt{Stack elec} and
\texttt{Stack ubuntu} datasets, the words \textit{item} and
\textit{review} will be changed to \textit{question} and
\textit{answer}. For the \texttt{ICEWS1819} and \texttt{GDELT}
datasets, words \textit{item} and \textit{user} will be unified as
\textit{entity}, and the word \textit{review} will be changed to
\textit{relation}. For the \texttt{Enron} dataset, words
\textit{item} and \textit{user} will be unified as \textit{user},
and the word \textit{review} will be changed to \textit{e-mail}.

\subsection{Additional Experiment Results}

\subsubsection{Edge Classification}
\label{appd:edge_classification_result}

\begin{figure}[t]
\centering
\subfigure[GDELT]{\includegraphics[width=0.32\linewidth]{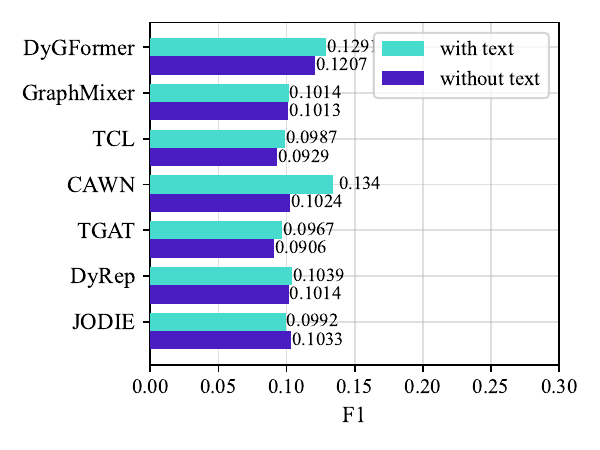}}
\subfigure[Stack elec]{\includegraphics[width=0.32\linewidth]{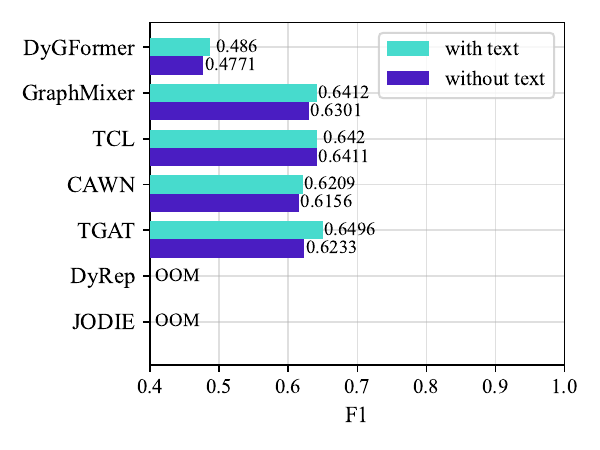}}
\subfigure[Stack ubuntu]{\includegraphics[width=0.32\linewidth]{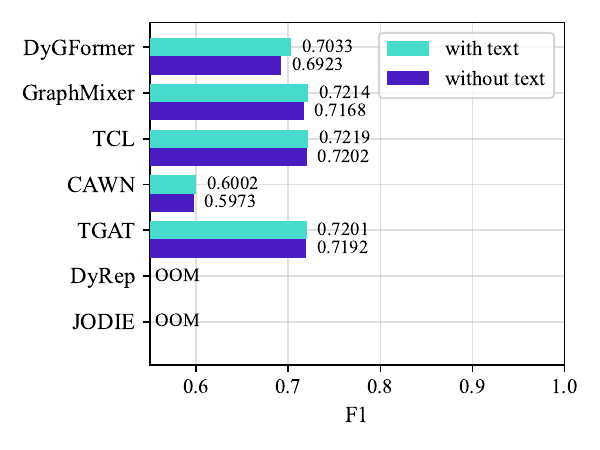}}
\subfigure[Amazon movies]{\includegraphics[width=0.32\linewidth]{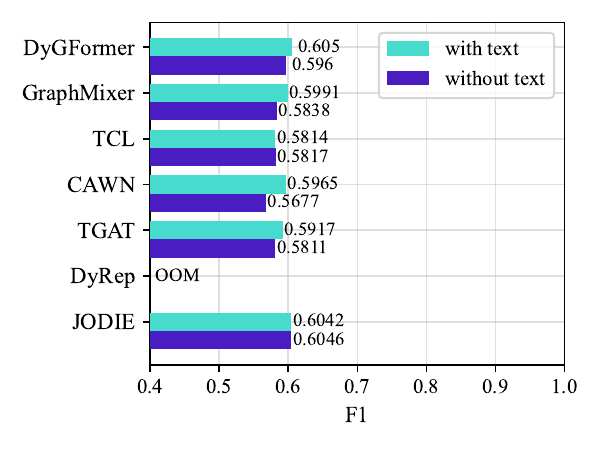}}
\subfigure[Yelp]{\includegraphics[width=0.32\linewidth]{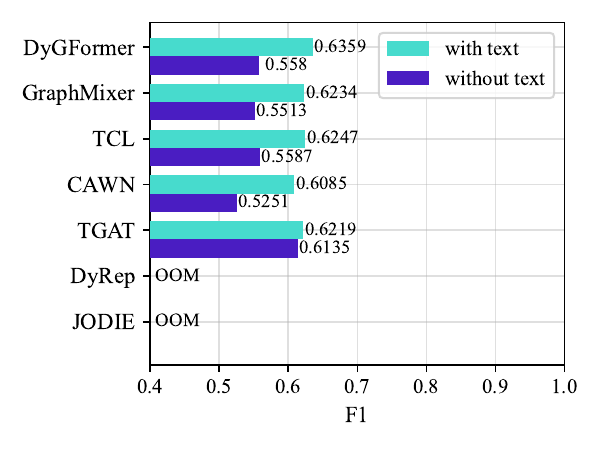}}
\caption{Edge classification performance with and without text attributes.}
\label{fig:edge_classification_bert_ablation_appendix}
\end{figure}

As illustrated in
Figure~\ref{fig:edge_classification_bert_ablation_appendix}, we can
see that text information can improve the edge classification
performance of existing models on datasets from different domains,
indicating the effectiveness of integrating edge text attributes in
discriminating the interaction types. However, some models have
performance degradation with text information, such as TCL on the
\texttt{Amazon movies} dataset, showing the weakness of some models
in handling rich semantic information. This observation demonstrates
the necessity of designing flexible methods to introduce text
information to existing dynamic graph learning methods.

\subsubsection{Future Link Prediction}
\label{appd:link_prediction_result}

As illustrated in Table~\ref{tab:link_prediction_appendix_AUC} and
Table~\ref{tab:link_prediction_appendix_AP}, we report the complete
results of the future link prediction task. We can see that the
memory-based models (\ie JODIE and DyRep) have scalability problems
facing large dynamic graphs (\eg \texttt{Stack ubuntu} and
\texttt{Yelp}). Since most dynamic graphs in the real world are
large and enriched with text attributes while many applications have
real-time requirements (\eg anomaly detection), it requires
lightweight models with accuracy and efficiency to adapt to
real-world scenarios. Furthermore, we can see that even though some
models get good performance in the future link prediction task, they
fail to achieve satisfactory performance in the destination node
retrieval task (\eg GraphMixer on the \texttt{Enron} dataset),
showing the limitations of existing models.

\subsubsection{Destination Node Retrieval}
\label{appd:node_retrieval_result}

\begin{figure}[t]
\centering
\subfigure[Enron]{\includegraphics[width=0.48\linewidth]{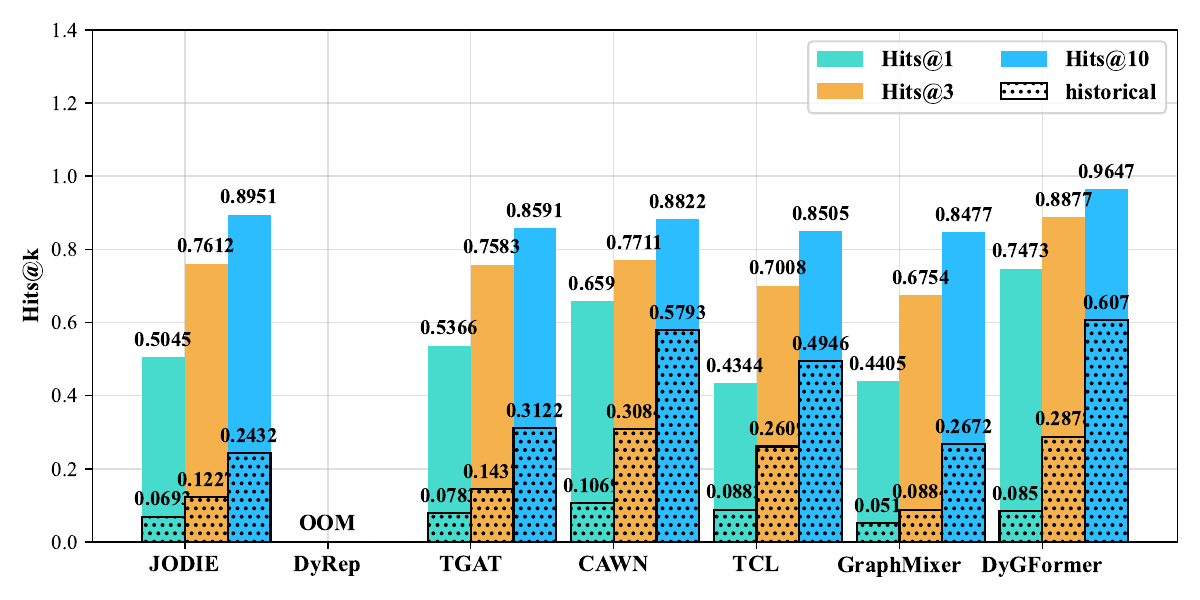}}
\subfigure[ICEWS1819]{\includegraphics[width=0.48\linewidth]{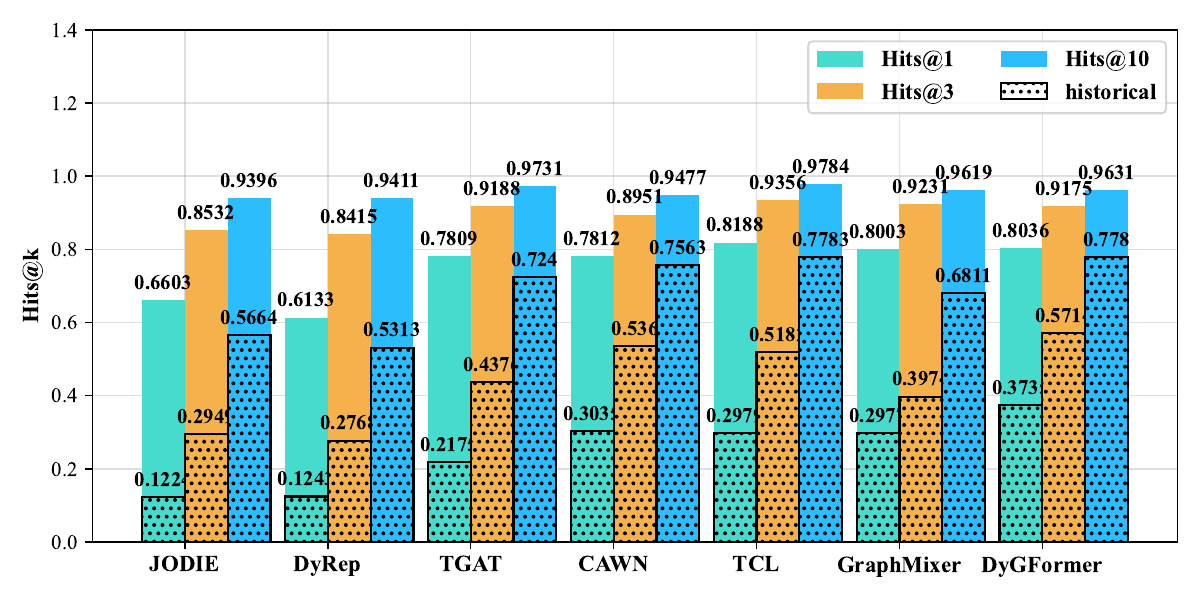}}
\caption{Node retrieval performance using random sampling and
historical sampling.}
\label{fig:node_retrieval_sample_ablation_appendix}
\end{figure}

As illustrated in Table~\ref{tab:node_retrieval_appendix_H1} and
Table~\ref{tab:node_retrieval_appendix_H3}, we provide more results
of the destination node retrieval task in the Hits@1 and Hits@3
metrics. As illustrated in
Figure~\ref{fig:node_retrieval_sample_ablation_appendix}, we provide
the ablation study result of sampling strategies on the
\texttt{Enron} and \texttt{ICEWS1819} datasets. The results meet our
observations in Section~\ref{sec:node_retrieval} and indicate the
limitations of existing models.

\begin{table}[t]
\caption{AUC-ROC for future link prediction. \textit{tr.} means
transductive setting and \textit{in.} means inductive setting.
\textbf{Text} means whether to use Bert-encoded embeddings for
initialization.} \centering \resizebox{1\textwidth}{!}{
\begin{tabular}{c|c|c|ccccccc}
\toprule
\textbf{}
&\textbf{Datasets}
&\textbf{Text}
&\multicolumn{1}{c}{\textbf{JODIE}}
&\multicolumn{1}{c}{\textbf{DyRep}}
&\multicolumn{1}{c}{\textbf{TGAT}}
&\multicolumn{1}{c}{\textbf{CAWN}}
&\multicolumn{1}{c}{\textbf{TCL}}
&\multicolumn{1}{c}{\textbf{GraphMixer}}
&\multicolumn{1}{c}{\textbf{DyGFormer}}\\
\midrule
&\multirow{2}{*}{\textbf{Enron}}  &\xmark &\textbf{0.9712 $\pm$ 0.0097} &0.9545 $\pm$ 0.0023 &0.9511 $\pm$ 0.0011 &0.9652 $\pm$ 0.0012 &0.9604 $\pm$ 0.0079 &0.9254 $\pm$ 0.0046 &0.9653 $\pm$ 0.0015\\

& &\cmark &0.9731 $\pm$ 0.0052 &0.9274 $\pm$ 0.0026 &0.9681 $\pm$ 0.0026 &0.9740 $\pm$ 0.0007 &0.9618 $\pm$ 0.0025 &0.9567 $\pm$ 0.0013 &\textbf{0.9779 $\pm$ 0.0014}\\

\cmidrule{2-10}

 &\multirow{2}{*}{\textbf{ICEWS1819}}&\xmark &0.9821 $\pm$ 0.0095 &0.9799 $\pm$ 0.0039 & 0.9787 $\pm$ 0.0065 &0.9815 $\pm$ 0.0041 &0.9842 $\pm$ 0.0036 &0.9399 $\pm$ 0.0079 &\textbf{0.9865 $\pm$ 0.0024}\\

 & &\cmark &0.9741 $\pm$ 0.0113 &0.9632 $\pm$ 0.0027 &0.9904 $\pm$ 0.0039 &0.9857  $\pm$ 0.0018 &\textbf{0.9923 $\pm$ 0.0012} &0.9863 $\pm$ 0.0024 &0.9888  $\pm$ 0.0015\\

 \cmidrule{2-10}

\textit{tr.} &\multirow{2}{*}{\textbf{Googlemap CT}} &\xmark &OOM &OOM &0.8537 $\pm$ 0.0153 &\textbf{0.8543 $\pm$ 0.0027} &0.7740 $\pm$ 0.0013 &0.7087 $\pm$ 0.0088 & 0.7864 $\pm$ 0.0047\\

 & &\cmark &OOM &OOM &\textbf{0.9049 $\pm$ 0.0071} &0.8687 $\pm$ 0.0063  &0.8348 $\pm$ 0.0094 &0.8095 $\pm$ 0.0014 &0.8207 $\pm$ 0.0018\\

\cmidrule{2-10}

&\multirow{2}{*}{\textbf{GDELT}} &\xmark &0.9562 $\pm$ 0.0027 &0.9477 $\pm$ 0.0011 &0.9341 $\pm$ 0.0046 &0.9419  $\pm$ 0.0026 &0.9571 $\pm$ 0.0007 &0.9316 $\pm$ 0.0021 &\textbf{0.9648 $\pm$ 0.0007}\\

 & &\cmark &0.9533 $\pm$ 0.0020 &0.9453 $\pm$ 0.0018 &0.9595 $\pm$ 0.0033 &0.9600 $\pm$ 0.0061 &0.9619  $\pm$ 0.0008 &0.9552 $\pm$ 0.0018 &\textbf{0.9662 $\pm$ 0.0003}\\

 \cmidrule{2-10}

 &\multirow{2}{*}{\textbf{Stack elec}} &\xmark & OOM & OOM & 0.9609 $\pm$ 0.0010 &0.9610 $\pm$ 0.0013 & 0.8796 $\pm$ 0.0089 & 0.9604 $\pm$ 0.0006 & \textbf{0.9788 $\pm$ 0.0018}\\
 & &\cmark & OOM & OOM &0.9709 $\pm$ 0.0014 &0.9631 $\pm$ 0.0007 &0.9578 $\pm$ 0.0106 &0.9673 $\pm$ 0.0011 &\textbf{0.9798 $\pm$ 0.0006} \\

 \cmidrule{2-10}

 &\multirow{2}{*}{\textbf{Stack ubuntu}} &\xmark &OOM &OOM &0.9180 $\pm$ 0.0520 &\textbf{0.9564 $\pm$ 0.0003} &0.9508 $\pm$ 0.0012 & 0.9522 $\pm$ 0.0006 &  0.9522 $\pm$ 0.0027 \\

 & &\cmark & OOM & OOM & 0.9490 $\pm$ 0.0018  & \textbf{0.9567 $\pm$ 0.0004} & 0.9516 $\pm$ 0.0003 & 0.9494 $\pm$ 0.0028  & 0.9526 $\pm$ 0.0035 \\

 \cmidrule{2-10}

 &\multirow{2}{*}{\textbf{Amazon movies}} &\xmark &OOM &OOM &0.8698 $\pm$ 0.0010 &0.8561 $\pm$ 0.0003 &0.8711 $\pm$ 0.0007 &\textbf{0.8726 $\pm$ 0.0003} &0.8713 $\pm$ 0.0005 \\

 & &\cmark &OOM &OOM &0.9064 $\pm$ 0.0014 &0.8936 $\pm$ 0.0012 &0.9050 $\pm$ 0.0012 &0.8894 $\pm$ 0.0008 &\textbf{0.9100 $\pm$ 0.0006}\\

 \cmidrule{2-10}

 &\multirow{2}{*}{\textbf{Yelp}} &\xmark &OOM &OOM &\textbf{0.8646 $\pm$ 0.0037} &0.8445 $\pm$ 0.0021 &0.8593 $\pm$ 0.0014 &0.8554 $\pm$ 0.0013 &0.8641 $\pm$ 0.0007 \\

 & &\cmark &OOM &OOM &0.9487 $\pm$ 0.0029 &0.9349 $\pm$ 0.0026 &\textbf{0.9528 $\pm$ 0.0018} &0.8927 $\pm$ 0.0021 &0.9407 $\pm$ 0.0010\\

 \midrule

 &\multirow{2}{*}{\textbf{Enron}}&\xmark &0.8745 $\pm$ 0.0041 &0.8560 $\pm$ 0.0124 &0.8079 $\pm$ 0.0047 &0.8710 $\pm$ 0.0030 &0.8363 $\pm$ 0.0068 &0.7510 $\pm$ 0.0071 &\textbf{0.8991 $\pm$ 0.0012}\\

 & &\cmark &0.8732 $\pm$ 0.0037 &0.7901 $\pm$ 0.0047 &0.8650 $\pm$ 0.0032 &0.9091 $\pm$ 0.0014 &0.8512 $\pm$ 0.0062 &0.8347 $\pm$ 0.0039 &\textbf{0.9316 $\pm$ 0.0015}\\

\cmidrule{2-10}

 &\multirow{2}{*}{\textbf{ICEWS1819}} &\xmark &0.9115 $\pm$ 0.0081 &0.9390  $\pm$ 0.0054 &0.9151 $\pm$ 0.0061 &0.9330  $\pm$ 0.0076 &0.9471 $\pm$ 0.0011 &0.8858 $\pm$ 0.0089 &\textbf{0.9613 $\pm$ 0.0010}\\

 & &\cmark &0.9285 $\pm$ 0.0065 &0.9030 $\pm$ 0.0097 &0.9706 $\pm$ 0.0054 &0.9774 $\pm$ 0.0039 &\textbf{0.9778 $\pm$ 0.0012}  &0.9605  $\pm$ 0.0025 &0.9630  $\pm$ 0.0027\\

 \cmidrule{2-10}

\textit{in.} &\multirow{2}{*}{\textbf{Googlemap CT}} &\xmark &OOM &OOM &0.7958 $\pm$ 0.0012 &\textbf{0.7968 $\pm$ 0.0007} &0.7104 $\pm$ 0.0015 &0.6675 $\pm$ 0.0033 &0.7148  $\pm$ 0.0024\\

 & &\cmark &OOM &OOM &\textbf{0.8791 $\pm$ 0.0028} &0.7058 $\pm$ 0.0047 &0.7895 $\pm$ 0.0046 &0.7543 $\pm$ 0.0018 & 0.7648 $\pm$ 0.0052\\

\cmidrule{2-10}

&\multirow{2}{*}{\textbf{GDELT}} &\xmark &0.8977 $\pm$ 0.0035 &0.8791 $\pm$ 0.0002 &0.7501 $\pm$ 0.0074 &0.7909  $\pm$ 0.0010  & 0.8544$\pm$ 0.0045 &0.7361 $\pm$ 0.0058 &\textbf{0.9135$\pm$ 0.0024}\\

& &\cmark &0.8921 $\pm$ 0.0065 &0.8917 $\pm$ 0.0007 &0.9012 $\pm$ 0.0011  &0.8899 $\pm$ 0.0082 &0.9099 $\pm$ 0.0022 &0.8942 $\pm$ 0.0035  &\textbf{0.9206 $\pm$ 0.0003}\\

\cmidrule{2-10}

&\multirow{2}{*}{\textbf{Stack elec}} &\xmark & OOM &OOM & 0.7800 $\pm$ 0.0024 &0.7833 $\pm$ 0.0089 &0.7243 $\pm$ 0.0074 & 0.7856 $\pm$ 0.0039 & \textbf{0.8578 $\pm$ 0.0035} \\

 & &\cmark &OOM &OOM &0.8423 $\pm$ 0.0018 &0.7963 $\pm$ 0.0074 &0.7689 $\pm$ 0.0051 &0.8232 $\pm$ 0.0031 &\textbf{0.8607 $\pm$ 0.0015} \\

 \cmidrule{2-10}

 &\multirow{2}{*}{\textbf{Stack ubuntu}} &\xmark &OOM &OOM & 0.7213 $\pm$ 0.0549 & 0.7892 $\pm$ 0.0039 &0.7712 $\pm$ 0.0036 &0.7752 $\pm$ 0.0031 & \textbf{0.7895 $\pm$ 0.0018}\\

 & &\cmark & OOM & OOM &0.7655 $\pm$ 0.0019 &\textbf{0.7871 $\pm$ 0.008}& 0.7717 $\pm$ 0.0027 &0.7870 $\pm$ 0.0032 &0.7773 $\pm$ 0.0047 \\
\cmidrule{2-10}

 &\multirow{2}{*}{\textbf{Amazon movies}} &\xmark &OOM &OOM &\textbf{0.8186 $\pm$ 0.0017} &0.7910 $\pm$ 0.0012 &0.8129 $\pm$ 0.0004 &0.8139 $\pm$ 0.0008 &0.8140 $\pm$ 0.0007\\

 & &\cmark &OOM &OOM &0.8706 $\pm$ 0.0023 &0.8492 $\pm$ 0.0008  &0.8677 $\pm$ 0.0009 &0.8418 $\pm$ 0.0012 &\textbf{0.8733 $\pm$ 0.0005} \\

\cmidrule{2-10}

 &\multirow{2}{*}{\textbf{Yelp}} &\xmark &OOM &OOM &\textbf{0.8090 $\pm$ 0.0011} &0.7843 $\pm$ 0.0018 &0.8011 $\pm$ 0.0009 &0.7951 $\pm$ 0.0006 &0.8058 $\pm$ 0.0005\\

 & &\cmark &OOM &OOM &0.9173 $\pm$ 0.0008 &0.8995 $\pm$ 0.0005 &\textbf{0.9233 $\pm$ 0.0011} &0.8452 $\pm$ 0.0014 &0.9067 $\pm$ 0.0009\\

\bottomrule
\end{tabular}}
\label{tab:link_prediction_appendix_AUC}
\end{table}

\begin{table}[t]
\caption{AP for future link prediction. \textit{tr.} means
transductive setting and \textit{in.} means inductive setting.
\textbf{Text} means whether to use Bert-encoded embeddings for
initialization.} \centering \resizebox{1\textwidth}{!}{
\begin{tabular}{c|c|c|ccccccc}
\toprule
\textbf{}
&\textbf{Datasets}
&\textbf{Text}
&\multicolumn{1}{c}{\textbf{JODIE}}
&\multicolumn{1}{c}{\textbf{DyRep}}
&\multicolumn{1}{c}{\textbf{TGAT}}
&\multicolumn{1}{c}{\textbf{CAWN}}
&\multicolumn{1}{c}{\textbf{TCL}}
&\multicolumn{1}{c}{\textbf{GraphMixer}}
&\multicolumn{1}{c}{\textbf{DyGFormer}}\\
\midrule
&\multirow{2}{*}{\textbf{Enron}}  &\xmark &0.9566 $\pm$ 0.0047 &0.9498 $\pm$ 0.0059 &0.9490 $\pm$ 0.0015 &0.9664 $\pm$ 0.0009 &0.9594 $\pm$ 0.0032 &0.9138 $\pm$ 0.0035 &\textbf{0.9716 $\pm$ 0.0018}\\

& &\cmark &0.9553 $\pm$ 0.0051 &0.9066 $\pm$ 0.0076 &0.9668 $\pm$ 0.0026 &0.9756 $\pm$ 0.0008 &0.9603 $\pm$ 0.0018 &0.9559 $\pm$ 0.0027 &\textbf{0.9804 $\pm$ 0.0015}\\

\cmidrule{2-10}

 &\multirow{2}{*}{\textbf{ICEWS1819}}&\xmark &0.9824 $\pm$ 0.0062 &0.9813 $\pm$ 0.0029 &0.9805 $\pm$ 0.0052 &0.9838 $\pm$ 0.0031 &0.9866 $\pm$ 0.0017 &0.9610 $\pm$ 0.0085 &\textbf{0.9884 $\pm$ 0.0013}\\

 & &\cmark &0.9752 $\pm$ 0.0037 &0.9676 $\pm$ 0.0026 &0.9908 $\pm$ 0.0032 &0.9886 $\pm$ 0.0025 &\textbf{0.9927  $\pm$ 0.0012} &0.9871 $\pm$ 0.0034 &0.9901 $\pm$ 0.0018\\

 \cmidrule{2-10}

\textit{tr.} &\multirow{2}{*}{\textbf{Googlemap CT}} &\xmark &OOM &OOM &0.8480 $\pm$ 0.0026 &\textbf{0.8511 $\pm$ 0.0022}  &0.7789 $\pm$ 0.0014 &0.6768 $\pm$ 0.0071 &0.7865 $\pm$ 0.0023\\

 & &\cmark &OOM &OOM &\textbf{0.9002 $\pm$ 0.0019} &0.8721 $\pm$ 0.0027 &0.8335 $\pm$ 0.0018 &0.8072 $\pm$ 0.0010 &0.8183 $\pm$ 0.0038 \\

\cmidrule{2-10}

&\multirow{2}{*}{\textbf{GDELT}} &\xmark &0.9482 $\pm$ 0.0018 &0.9415 $\pm$ 0.0013 &0.9342 $\pm$ 0.0039 &0.9398 $\pm$ 0.0036 &0.9554 $\pm$ 0.0006 &0.9299 $\pm$ 0.0028 &\textbf{0.9640 $\pm$ 0.0002}\\

 & &\cmark &0.9466 $\pm$ 0.0032 &0.9416 $\pm$ 0.0017 &0.9572 $\pm$ 0.0029 &0.9582 $\pm$ 0.0053 &0.9601 $\pm$ 0.0011 &0.9523 $\pm$ 0.0020 & \textbf{0.9653 $\pm$ 0.0003}\\

 \cmidrule{2-10}

 &\multirow{2}{*}{\textbf{Stack elec}} &\xmark &OOM &OOM &0.9489 $\pm$ 0.0003 &0.9496 $\pm$ 0.0031 & 0.8613 $\pm$ 0.0062 & 0.9502 $\pm$ 0.0011 & \textbf{0.9746 $\pm$ 0.0015}\\
 & &\cmark &OOM &OOM &0.9646 $\pm$ 0.0005 &0.9529 $\pm$ 0.0023 &0.9441 $\pm$ 0.0079 & 0.9591 $\pm$ 0.0009 &\textbf{0.9819 $\pm$ 0.0010} \\

 \cmidrule{2-10}

 &\multirow{2}{*}{\textbf{Stack ubuntu}} &\xmark &OOM &OOM & 0.9040 $\pm$ 0.0564 &\textbf{0.9481 $\pm$ 0.0007} &0.9412 $\pm$ 0.0018  & 0.9417 $\pm$ 0.0016 & 0.9432 $\pm$ 0.0026\\

 & &\cmark &OOM &OOM &0.9352 $\pm$ 0.0012 &0.9420 $\pm$ 0.0005 & 0.9416$\pm$ 0.0023 & 0.9416$\pm$ 0.0047 & \textbf{0.9431$\pm$ 0.0008} \\

 \cmidrule{2-10}

 &\multirow{2}{*}{\textbf{Amazon movies}} &\xmark &OOM &OOM &0.8733 $\pm$ 0.0011 &0.8502 $\pm$ 0.0003 &0.8762 $\pm$ 0.0005 &\textbf{0.8776 $\pm$ 0.0002} &0.8766 $\pm$ 0.0007 \\

 & &\cmark &OOM &OOM &0.9065 $\pm$ 0.0016 &0.8885 $\pm$ 0.0007 &0.9051 $\pm$ 0.0010 &0.8906 $\pm$ 0.0006 &\textbf{0.9097 $\pm$ 0.0003} \\

 \cmidrule{2-10}

 &\multirow{2}{*}{\textbf{Yelp}} &\xmark &OOM &OOM &0.8622 $\pm$ 0.0032 &0.8396 $\pm$ 0.0018 &0.8583 $\pm$ 0.0017 &0.8472 $\pm$ 0.0010 &\textbf{0.8679 $\pm$ 0.0008} \\

 & &\cmark &OOM &OOM &0.9457 $\pm$ 0.0025 &0.9323 $\pm$ 0.0027 &\textbf{0.9511 $\pm$ 0.0013} &0.8883 $\pm$ 0.0026 &0.9391 $\pm$ 0.0011\\

 \midrule

 &\multirow{2}{*}{\textbf{Enron}}&\xmark &0.8635 $\pm$ 0.0018 &0.8682 $\pm$ 0.0049 &0.8180 $\pm$ 0.0034 &0.8901 $\pm$ 0.0023 &0.8433 $\pm$ 0.0043 &0.7482 $\pm$ 0.0066 &\textbf{0.9199 $\pm$ 0.0021}\\

 & &\cmark &0.8761 $\pm$ 0.0023 &0.7734 $\pm$ 0.0044 &0.8589 $\pm$ 0.0031 &0.9223 $\pm$ 0.0011 &0.8560 $\pm$ 0.0024 &0.8328 $\pm$ 0.0034 &\textbf{0.9409 $\pm$ 0.0025}\\

\cmidrule{2-10}

 &\multirow{2}{*}{\textbf{ICEWS1819}} &\xmark &0.9382 $\pm$ 0.0071 &0.9444 $\pm$ 0.0037 &0.9181 $\pm$ 0.0056 &0.9406 $\pm$ 0.0028 &0.9536 $\pm$ 0.0018 &0.8811 $\pm$ 0.0076 &\textbf{0.9665 $\pm$ 0.0011}\\

 & &\cmark &0.9333 $\pm$ 0.0026 &0.9134 $\pm$ 0.0041 &0.9716 $\pm$ 0.0033 &0.9631 $\pm$ 0.0034 &\textbf{0.9789 $\pm$ 0.0022} &0.9625 $\pm$ 0.0030 &0.9688 $\pm$ 0.0018\\

 \cmidrule{2-10}

\textit{in.} &\multirow{2}{*}{\textbf{Googlemap CT}} &\xmark &OOM &OOM &0.7843 $\pm$ 0.0020 &\textbf{0.7942 $\pm$ 0.0024} &0.7274 $\pm$ 0.0016 &0.6624 $\pm$ 0.0060 &0.7293 $\pm$ 0.0011\\

 & &\cmark &OOM &OOM &\textbf{0.8750 $\pm$ 0.0015} &0.8012 $\pm$ 0.0021 &0.7936 $\pm$ 0.0009 &0.7633 $\pm$ 0.0013 &0.7735 $\pm$ 0.0031 \\

\cmidrule{2-10}

&\multirow{2}{*}{\textbf{GDELT}} &\xmark &0.9072 $\pm$ 0.0017 &0.8756 $\pm$ 0.0022 &0.7500 $\pm$ 0.0067 &0.7980 $\pm$ 0.0016 &0.8430  $\pm$ 0.0053 &0.7298 $\pm$ 0.0060 &\textbf{0.9172 $\pm$ 0.0014}\\

& &\cmark &0.9019 $\pm$ 0.0023 &0.8928 $\pm$ 0.0011 &0.9023 $\pm$ 0.0010 &0.8986 $\pm$ 0.0077 &0.9151 $\pm$ 0.0045 &0.8925 $\pm$ 0.0048 &\textbf{0.9263 $\pm$ 0.0009}\\

\cmidrule{2-10}

&\multirow{2}{*}{\textbf{Stack elec}} &\xmark & OOM & OOM & 0.7784 $\pm$ 0.0043 & 0.7816 $\pm$ 0.0123 &0.7346 $\pm$ 0.0068 & 0.7874 $\pm$ 0.0040 & \textbf{0.8742 $\pm$ 0.0061} \\

 & &\cmark &OOM &OOM &0.8391 $\pm$ 0.0036 &0.7921$\pm$ 0.0086 &0.7599 $\pm$ 0.0041 &0.8142 $\pm$ 0.0021 &\textbf{0.8801 $\pm$ 0.0043}\\

 \cmidrule{2-10}

 &\multirow{2}{*}{\textbf{Stack ubuntu}} &\xmark &OOM &OOM &0.7365 $\pm$ 0.0421 &0.7957 $\pm$ 0.0030 &0.7746 $\pm$ 0.0058 & 0.7764 $\pm$ 0.0105 & \textbf{0.8054 $\pm$ 0.0027} \\

 & &\cmark &OOM &OOM &0.7664 $\pm$ 0.0015 & \textbf{0.7882 $\pm$ 0.0015} & 0.7777 $\pm$ 0.0015 & 0.7870 $\pm$ 0.0015 & 0.7832 $\pm$ 0.0015 \\

\cmidrule{2-10}

 &\multirow{2}{*}{\textbf{Amazon movies}} &\xmark &OOM &OOM &0.8322 $\pm$ 0.0013 &0.7974 $\pm$ 0.0004 &0.8311 $\pm$ 0.0002 &\textbf{0.8325 $\pm$ 0.0003} &0.8324 $\pm$ 0.0004 \\

 & &\cmark &OOM &OOM &0.8760 $\pm$ 0.0010 &0.8508 $\pm$ 0.0006 &0.8738 $\pm$ 0.005 &0.8517 $\pm$ 0.0007 &\textbf{0.8780 $\pm$ 0.0006} \\

\cmidrule{2-10}

 &\multirow{2}{*}{\textbf{Yelp}} &\xmark &OOM &OOM &0.8212 $\pm$ 0.0019 &0.7942 $\pm$ 0.0007 &0.8159 $\pm$ 0.0014 &0.8028 $\pm$ 0.0008 &\textbf{0.8251 $\pm$ 0.0015} \\

 & &\cmark &OOM &OOM &0.9174 $\pm$ 0.0010 &0.9010 $\pm$ 0.0009 &\textbf{0.9249 $\pm$ 0.0013} &0.8494 $\pm$ 0.0008 &0.9092 $\pm$ 0.0006 \\

\bottomrule
\end{tabular}}
\label{tab:link_prediction_appendix_AP}
\end{table}

\begin{table}[t]
\caption{Hits@1 for destination node retrieval. \textit{tr.} means
transductive setting and \textit{in.} means inductive setting.
\textbf{Text} means whether to use Bert-encoded embeddings for
initialization.} \centering \resizebox{1\textwidth}{!}{
\begin{tabular}{c|c|c|ccccccc}
\toprule
\textbf{}
&\textbf{Datasets}
&\textbf{Text}
&\multicolumn{1}{c}{\textbf{JODIE}}
&\multicolumn{1}{c}{\textbf{DyRep}}
&\multicolumn{1}{c}{\textbf{TGAT}}
&\multicolumn{1}{c}{\textbf{CAWN}}
&\multicolumn{1}{c}{\textbf{TCL}}
&\multicolumn{1}{c}{\textbf{GraphMixer}}
&\multicolumn{1}{c}{\textbf{DyGFormer}}\\
\midrule
&\multirow{2}{*}{\textbf{Enron}}  &\xmark &0.4610 $\pm$ 0.0078 &OOM &0.2649 $\pm$ 0.0039 &0.0530 $\pm$ 0.0041 &0.3083 $\pm$ 0.0023 &0.3107 $\pm$ 0.0022 &\textbf{0.4928 $\pm$ 0.0011} \\

& &\cmark &0.5045 $\pm$ 0.0083 &OOM &0.5366 $\pm$ 0.0026 &0.6590 $\pm$ 0.0018 &0.4344 $\pm$ 0.0020 &0.4405 $\pm$ 0.0015 &\textbf{0.7473 $\pm$ 0.0023}\\

\cmidrule{2-10}

 &\multirow{2}{*}{\textbf{ICEWS1819}}&\xmark &0.6437 $\pm$ 0.0017 &0.6329 $\pm$ 0.0024 &0.4523 $\pm$ 0.0053 &0.4755 $\pm$ 0.0049 &0.5142 $\pm$ 0.0033 &0.5288 $\pm$ 0.0016 &\textbf{0.6630 $\pm$ 0.0013}\\

 & &\cmark  &0.6603 $\pm$ 0.0010   &0.6133 $\pm$ 0.0026 &0.7809 $\pm$ 0.0031 &0.7812 $\pm$ 0.0051   &\textbf{0.8188 $\pm$ 0.0121} &0.8003 $\pm$ 0.0078    &0.8036 $\pm$ 0.0002\\

 \cmidrule{2-10}

\textit{tr.} &\multirow{2}{*}{\textbf{Googlemap CT}} &\xmark &OOM &OOM  &0.0127 $\pm$ 0.0065  &0.0122 $\pm$ 0.0034    &\textbf{0.0156 $\pm$ 0.0002}  &0.0110 $\pm$ 0.0001 &0.0133 $\pm$ 0.0005\\

 & &\cmark &OOM &OOM &\textbf{0.2290 $\pm$ 0.0042} &0.1322 $\pm$ 0.0019 &0.1468 $\pm$ 0.0004 &0.1202 $\pm$ 0.0005 &0.1314 $\pm$ 0.0002\\

\cmidrule{2-10}

&\multirow{2}{*}{\textbf{GDELT}} &\xmark &0.3323 $\pm$ 0.0027 &0.3127 $\pm$ 0.0015 &0.2453 $\pm$ 0.0017 &0.2177 $\pm$ 0.0082 &0.2083 $\pm$ 0.0037 &0.0611 $\pm$ 0.0052 &\textbf{0.3556 $\pm$ 0.0029}\\

 & &\cmark &0.2890 $\pm$ 0.0031 &0.3112 $\pm$ 0.0029 &0.4135 $\pm$ 0.0013 &0.4253 $\pm$ 0.0010 &0.4353 $\pm$ 0.0024 &0.3875 $\pm$ 0.0033 &\textbf{0.4764 $\pm$ 0.0008}\\

 \cmidrule{2-10}

%&\multirow{2}{*}{\textbf{Stack elec}} &\xmark &- &- &- &- &- &- &-\\
%& &\cmark &- &- &- &- &- &- &-\\
%\cmidrule{2-10}

% &\multirow{2}{*}{\textbf{Stack ubuntu}} &\xmark &- &- &- &- &- &- &-\\
% & &\cmark &- &- &- &- &- &- &-\\
%\cmidrule{2-10}

 &\multirow{2}{*}{\textbf{Amazon movies}} &\xmark &OOM &OOM &0.2407 $\pm$ 0.0013   &0.1678 $\pm$ 0.0023 &0.2373 $\pm$ 0.0014 &\textbf{0.2504 $\pm$ 0.0007} &0.2139 $\pm$ 0.0009 \\

 & &\cmark &OOM &OOM &0.2851 $\pm$ 0.0009 &0.2237 $\pm$ 0.0026 &0.2841 $\pm$ 0.0013 &0.2613 $\pm$ 0.0008    &\textbf{0.3008 $\pm$ 0.0007}\\

 \cmidrule{2-10}

 &\multirow{2}{*}{\textbf{Yelp}} &\xmark &OOM &OOM &\textbf{0.1774 $\pm$ 0.0039}   &0.1473 $\pm$ 0.0024 &0.1747 $\pm$ 0.0038 &0.1554 $\pm$ 0.0031   &0.1284 $\pm$ 0.0017\\

 & &\cmark &OOM &OOM &\textbf{0.3737 $\pm$ 0.0051} &0.3218 $\pm$ 0.0104 &0.1554 $\pm$ 0.0028 &0.2204 $\pm$ 0.0037 &0.3512 $\pm$ 0.0027\\

 \midrule

 &\multirow{2}{*}{\textbf{Enron}}&\xmark &0.2719  $\pm$ 0.0055 &OOM &0.1301 $\pm$ 0.0027 &0.0391 $\pm$ 0.0016 &0.1473 $\pm$ 0.0025 &0.1518 $\pm$ 0.0013 &\textbf{0.2964 $\pm$ 0.0019} \\

 & &\cmark &0.3003 $\pm$ 0.0042 &OOM &0.2832 $\pm$ 0.0033 &0.4223 $\pm$ 0.0029 &0.1967 $\pm$ 0.0014 &0.2046 $\pm$ 0.0009 &\textbf{0.5553 $\pm$ 0.0027}\\

\cmidrule{2-10}

 &\multirow{2}{*}{\textbf{ICEWS1819}} &\xmark &0.4834 $\pm$ 0.0126 &0.4657 $\pm$ 0.0084 &0.2209 $\pm$ 0.0076 &0.4077 $\pm$ 0.0008 &0.2883 $\pm$ 0.0145 &0.2979 $\pm$ 0.0023 &\textbf{0.5009 $\pm$ 0.0073}\\

 & &\cmark &0.5134 $\pm$ 0.0010 &0.4799 $\pm$ 0.0026 &0.5752 $\pm$ 0.0031 &0.6342 $\pm$ 0.0051 &0.6059 $\pm$ 0.0122 &0.6114 $\pm$ 0.0078 &\textbf{0.634 $\pm$ 0.00028}\\

 \cmidrule{2-10}

\textit{in.} &\multirow{2}{*}{\textbf{Googlemap CT}} &\xmark &OOM &OOM &0.0109 $\pm$ 0.0007    &0.0105 $\pm$ 0.0003 &0.0118 $\pm$ 0.0005 &\textbf{0.0170 $\pm$ 0.0001} &0.0055 $\pm$ 0.0002\\

 & &\cmark &OOM &OOM &\textbf{0.0550 $\pm$ 0.0007} &0.0321 $\pm$ 0.0004 &0.0236 $\pm$ 0.0005 &0.0191 $\pm$ 0.0001 &0.0178 $\pm$ 0.0006\\

\cmidrule{2-10}

&\multirow{2}{*}{\textbf{GDELT}} &\xmark &\textbf{0.3040 $\pm$ 0.0061} &0.2678 $\pm$ 0.0088 &0.1270 $\pm$ 0.0056 &0.2138 $\pm$ 0.0043 &0.1854 $\pm$ 0.0091 &0.0241 $\pm$ 0.0036 &0.2092 $\pm$ 0.0027\\

& &\cmark &0.3080 $\pm$ 0.0074 &0.2740 $\pm$ 0.0051 &0.2630 $\pm$ 0.0042 &0.2916 $\pm$ 0.0011 &0.3199 $\pm$ 0.0031 &0.2610 $\pm$ 0.0046 &\textbf{0.3681 $\pm$ 0.0038}\\

\cmidrule{2-10}

%&\multirow{2}{*}{\textbf{Stack elec}} &\xmark &- &- &- &- &- &- &-\\
%& &\cmark &- &- &- &- &- &- &-\\
%\cmidrule{2-10}

% &\multirow{2}{*}{\textbf{Stack ubuntu}} &\xmark &- &- &- &- &- &- &-\\
% & &\cmark &- &- &- &- &- &- &-\\
%\cmidrule{2-10}

 &\multirow{2}{*}{\textbf{Amazon movies}} &\xmark &OOM &OOM &\textbf{0.2087 $\pm$ 0.0037}  &0.1406 $\pm$ 0.0017 &0.2071 $\pm$ 0.0015 &0.1906 $\pm$ 0.0004 &0.1886 $\pm$ 0.0038 \\

 & &\cmark &OOM &OOM &0.2403 $\pm$ 0.0038 &0.1835 $\pm$ 0.0042 &0.2463 $\pm$ 0.0008  &0.2224 $\pm$ 0.0009 &\textbf{0.2613 $\pm$ 0.0003}\\

\cmidrule{2-10}

 &\multirow{2}{*}{\textbf{Yelp}} &\xmark &OOM &OOM &\textbf{0.1515 $\pm$ 0.0105}   &0.1240 $\pm$ 0.0087   &0.1490 $\pm$ 0.0108 &0.1383 $\pm$ 0.0054 &0.0984 $\pm$ 0.0042\\

 & &\cmark &OOM &OOM &\textbf{0.3086 $\pm$ 0.0238} &0.2642 $\pm$ 0.0071 &0.1383 $\pm$ 0.0105 &0.1841 $\pm$ 0.0014 &0.2898 $\pm$ 0.0036\\

\bottomrule
\end{tabular}}
\label{tab:node_retrieval_appendix_H1}
\end{table}

\begin{table}[t]
\caption{Hits@3 for destination node retrieval. \textit{tr.} means
transductive setting and \textit{in.} means inductive setting.
\textbf{Text} means whether to use Bert-encoded embeddings for
initialization.} \centering \resizebox{1\textwidth}{!}{
\begin{tabular}{c|c|c|ccccccc}
\toprule \textbf{} &\textbf{Datasets} &\textbf{Text}
&\multicolumn{1}{c}{\textbf{JODIE}}
&\multicolumn{1}{c}{\textbf{DyRep}}
&\multicolumn{1}{c}{\textbf{TGAT}}
&\multicolumn{1}{c}{\textbf{CAWN}} &\multicolumn{1}{c}{\textbf{TCL}}
&\multicolumn{1}{c}{\textbf{GraphMixer}}
&\multicolumn{1}{c}{\textbf{DyGFormer}}\\
\midrule
&\multirow{2}{*}{\textbf{Enron}}  &\xmark &\textbf{0.7297 $\pm$ 0.0412} &OOM &0.5103 $\pm$ 0.0023 &0.3084 $\pm$ 0.0037 &0.5085 $\pm$ 0.0049 &0.5105 $\pm$ 0.0062 &0.6645 $\pm$ 0.0041 \\

& &\cmark &0.7612 $\pm$ 0.0311 &OOM &0.7583 $\pm$ 0.0082 &0.7711 $\pm$ 0.0064 &0.7008 $\pm$ 0.0081 &0.6754 $\pm$ 0.0027 &\textbf{0.8877 $\pm$ 0.0051}\\

\cmidrule{2-10}

 &\multirow{2}{*}{\textbf{ICEWS1819}}&\xmark &0.6437 $\pm$ 0.0024 &0.6016 $\pm$ 0.0018 &0.6512 $\pm$ 0.0027 &0.6836 $\pm$ 0.0019 &0.7111 $\pm$ 0.0039 &0.7094 $\pm$ 0.0046 &\textbf{0.8192 $\pm$ 0.0021}\\

 & &\cmark &0.8532 $\pm$ 0.0012    &0.8415 $\pm$ 0.0061 &0.9188 $\pm$ 0.0034 &0.8951 $\pm$ 0.0078   &\textbf{0.9356 $\pm$ 0.0106}  &0.9231 $\pm$ 0.0013   &0.9175 $\pm$ 0.0006\\

 \cmidrule{2-10}

\textit{tr.} &\multirow{2}{*}{\textbf{Googlemap CT}} &\xmark &OOM &OOM &\textbf{0.2579 $\pm$ 0.0058}   &0.1824 $\pm$ 0.0007 &0.1750 $\pm$ 0.0004 &0.1743 $\pm$ 0.0004 &0.2556 $\pm$ 0.0008\\

 & &\cmark &OOM &OOM &\textbf{0.4325 $\pm$ 0.0023} &0.2771 $\pm$ 0.0018 &0.2966$\pm$ 0.0031 &0.2514$\pm$ 0.0028 &0.2705$\pm$ 0.0021 \\

\cmidrule{2-10}

&\multirow{2}{*}{\textbf{GDELT}} &\xmark &\textbf{0.6184 $\pm$ 0.0042} &0.5694 $\pm$ 0.0067 &0.3052 $\pm$ 0.0072 &0.2649 $\pm$ 0.0053 &0.4560 $\pm$ 0.0080 &0.1585 $\pm$ 0.0076 &0.6130 $\pm$ 0.0013\\

 & &\cmark &0.5752 $\pm$ 0.0035 &0.5731 $\pm$ 0.0028 &0.6621 $\pm$ 0.0019 &0.6631 $\pm$ 0.0017 &0.6756 $\pm$ 0.0029 &0.6351 $\pm$ 0.0030 &\textbf{0.7089 $\pm$ 0.0010}\\

 \cmidrule{2-10}

%&\multirow{2}{*}{\textbf{Stack elec}} &\xmark &- &- &- &- &- &- &-\\
%& &\cmark &- &- &- &- &- &- &-\\
%\cmidrule{2-10}

% &\multirow{2}{*}{\textbf{Stack ubuntu}} &\xmark &- &- &- &- &- &- &-\\
% & &\cmark &- &- &- &- &- &- &-\\
%\cmidrule{2-10}

 &\multirow{2}{*}{\textbf{Amazon movies}} &\xmark &OOM &OOM &0.4128 $\pm$ 0.0038   &0.3189 $\pm$ 0.0039 &0.4133 $\pm$ 0.0053 &\textbf{0.4199 $\pm$ 0.0013} &0.3506 $\pm$ 0.0013 \\

 & &\cmark &OOM &OOM &0.4855 $\pm$ 0.0013 &0.4122 $\pm$ 0.0065 &0.4800 $\pm$ 0.0011  &0.4438 $\pm$ 0.0013 &\textbf{0.4990 $\pm$ 0.0008} \\

 \cmidrule{2-10}

 &\multirow{2}{*}{\textbf{Yelp}} &\xmark &OOM &OOM &\textbf{0.3471 $\pm$ 0.0064}   &0.2876 $\pm$ 0.0032 &0.3457 $\pm$ 0.0126 &0.2986 $\pm$ 0.0128   &0.2686 $\pm$ 0.0103\\

 & &\cmark &OOM &OOM &\textbf{0.5968 $\pm$ 0.0043} &0.5393 $\pm$ 0.0024 &0.2986 $\pm$ 0.0084 &0.4017 $\pm$ 0.0042 &0.5549 $\pm$ 0.0054\\

 \midrule

 &\multirow{2}{*}{\textbf{Enron}}&\xmark &\textbf{0.4987 $\pm$ 0.0029} &OOM &0.2635 $\pm$ 0.0035 &0.2855 $\pm$ 0.0042 &0.2724 $\pm$ 0.0047 &0.2691 $\pm$ 0.0019 &0.4608 $\pm$ 0.0082 \\

 & &\cmark &0.5244 $\pm$ 0.0016 &OOM &0.5045 $\pm$ 0.0025 &0.5962 $\pm$ 0.0027 &0.3928 $\pm$ 0.0035 &0.3916 $\pm$ 0.0011 &\textbf{0.7474 $\pm$ 0.0026}\\

\cmidrule{2-10}

 &\multirow{2}{*}{\textbf{ICEWS1819}} &\xmark &\textbf{0.7035 $\pm$ 0.0125} &0.6673 $\pm$ 0.0193 &0.4148 $\pm$ 0.0032 &0.4723 $\pm$ 0.0079 &0.4663 $\pm$ 0.0128 &0.4417 $\pm$ 0.0113 &0.6748 $\pm$ 0.0087\\

 & &\cmark &0.7140 $\pm$ 0.0086 &0.7010 $\pm$ 0.0032 &0.7835 $\pm$ 0.0029 &\textbf{0.8120 $\pm$ 0.0066} &0.8026 $\pm$ 0.0091 &0.8001 $\pm$ 0.0088 &0.8017 $\pm$ 0.0071\\

 \cmidrule{2-10}

\textit{in.} &\multirow{2}{*}{\textbf{Googlemap CT}} &\xmark &OOM &OOM &0.0308 $\pm$ 0.0003    &0.0292 $\pm$ 0.0009 &\textbf{0.0352 $\pm$ 0.0012}    &0.0346 $\pm$ 0.0001 &0.0174 $\pm$ 0.0016\\

 & &\cmark &OOM &OOM &\textbf{0.1545 $\pm$ 0.0017} &0.0828 $\pm$ 0.0021 &0.0621 $\pm$ 0.0008 &0.0465 $\pm$ 0.0004 &0.0448 $\pm$ 0.0007 \\

\cmidrule{2-10}

&\multirow{2}{*}{\textbf{GDELT}} &\xmark &\textbf{0.5352 $\pm$ 0.0011} &0.4940 $\pm$ 0.0059 &0.2632 $\pm$ 0.0072 &0.2929 $\pm$ 0.0043 &0.3434 $\pm$ 0.0011 &0.0699 $\pm$ 0.0077 &0.4341 $\pm$ 0.0039 \\

& &\cmark &0.5294 $\pm$ 0.0028 &0.5179 $\pm$ 0.0068 &0.4818 $\pm$ 0.0055 &0.4667 $\pm$ 0.0061 &0.5090 $\pm$ 0.0024 &0.4755 $\pm$ 0.0081 &\textbf{0.5666 $\pm$ 0.0026}\\

\cmidrule{2-10}

%&\multirow{2}{*}{\textbf{Stack elec}} &\xmark &- &- &- &- &- &- &-\\
%& &\cmark &- &- &- &- &- &- &-\\
%\cmidrule{2-10}

% &\multirow{2}{*}{\textbf{Stack ubuntu}} &\xmark &- &- &- &- &- &- &-\\
% & &\cmark &- &- &- &- &- &- &-\\
%\cmidrule{2-10}

 &\multirow{2}{*}{\textbf{Amazon movies}} &\xmark &OOM &OOM &0.3511 $\pm$ 0.0087   &0.2634 $\pm$ 0.0033 &0.3496 $\pm$ 0.0028 &\textbf{0.3563 $\pm$ 0.0026} &0.3112 $\pm$ 0.0079 \\

 & &\cmark &OOM &OOM &0.4165 $\pm$ 0.0061 &0.3441 $\pm$ 0.0051 &0.4157 $\pm$ 0.0023  &0.3764 $\pm$ 0.0011 &\textbf{0.4319 $\pm$ 0.0011} \\

\cmidrule{2-10}

 &\multirow{2}{*}{\textbf{Yelp}} &\xmark &OOM &OOM &\textbf{0.3009 $\pm$ 0.0241}   &0.2420 $\pm$  0.0064  &0.2958 $\pm$ 0.0154 &0.2594 $\pm$ 0.0069 &0.2157 $\pm$ 0.0051\\

 & &\cmark &OOM &OOM &\textbf{0.5120 $\pm$ 0.0206} &0.4553 $\pm$ 0.0120 &0.2594 $\pm$ 0.0168 &0.3413 $\pm$ 0.0060 &0.4751 $\pm$ 0.0064\\

\bottomrule
\end{tabular}}
\label{tab:node_retrieval_appendix_H3}
\end{table}

\subsubsection{Textual Relation Generation}
\label{appd:relation_generation_result}

\begin{table}[t]
\centering \caption{Precision, Recall and F1 of BERTscore of
different LLMs for the textural relation generation tasks. The
number of test samples is 500 per
dataset.}\label{tab:text_generation_appendix}
\resizebox{0.8\linewidth}{!}{
\begin{tabular}{l|ccc|ccc|ccc}\toprule
 & \multicolumn{3}{c|}{\textbf{ICEWS1819}} & \multicolumn{3}{c|}{\textbf{Enron}} & \multicolumn{3}{c}{\textbf{Stack ubuntu}} \\
 & \multicolumn{1}{c}{Precision} & \multicolumn{1}{c}{Recall} & \multicolumn{1}{c|}{F1} & \multicolumn{1}{c}{Precision} & \multicolumn{1}{c}{Recall} & \multicolumn{1}{c|}{F1} & \multicolumn{1}{c}{Precision} & \multicolumn{1}{c}{Recall} & \multicolumn{1}{c}{F1} \\ \midrule
Llama3-8b & 77.38 & 82.23 & 79.71 & 79.88 & 79.26 & 79.46 & 79.48 & 82.86 & 81.10 \\
Mistral-7b &78.15 & \textbf{82.52} & 80.25 & 79.75 & \textbf{79.37} & 79.52  & 79.61 & 82.98 & 81.24\\
Vicuna-7b & 77.44 & 82.17 & 79.71 & \textbf{81.50} & 78.62 & \textbf{79.98}  & 80.92 & \textbf{83.01} & 81.93\\
Vicuna-13b & \textbf{78.81} & 82.38 & \textbf{80.51} & 80.74 & 78.99 & 79.81 & \textbf{81.39} & 82.99 & \textbf{82.15}\\
\bottomrule
\end{tabular}}
\end{table}

\begin{figure}[!h]
\centering \subfigure[Googlemap
CT]{\includegraphics[width=0.42\linewidth]{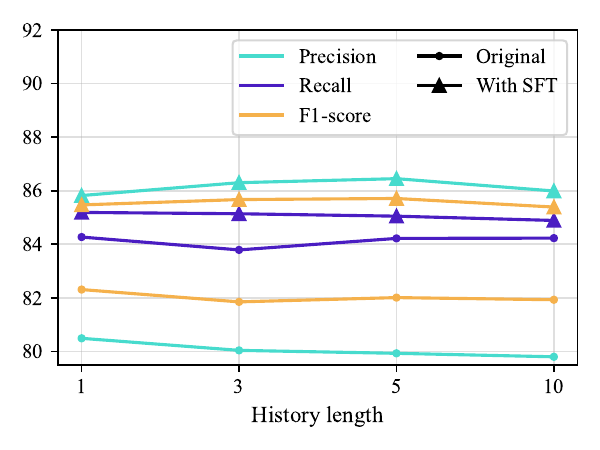}}
\subfigure[Stack
elec]{\includegraphics[width=0.42\linewidth]{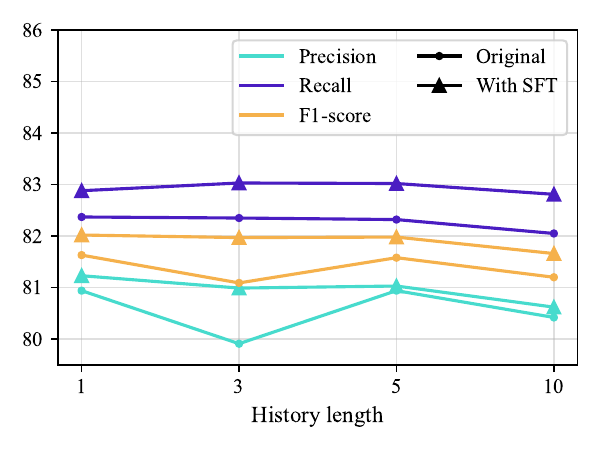}}
\caption{Textual relation generation performance with different
history lengths.}
\label{fig:text_generation_length_ablation_appendix}
\end{figure}

As shown in Table~\ref{tab:text_generation_appendix}, we provide the
performance of textual relation generation task on more datasets. As
shown in Figure~\ref{fig:text_generation_length_ablation_appendix},
the influence of the input history length is studied for the
relation generation performance. We report the performance of
Vicuna-7b on the \texttt{Googlemap CT} dataset and the performance
of Llama3-8b on the \texttt{Stack elec} dataset. We control the
history length by sampling the recent $k$ reviews. We have two
observations. First, the performance may degrade with a larger
history length (\eg precision on \texttt{Stack elec}). This
observation shows the necessity of designing strategies to flexibly
handle history text for different samples. Second, the fine-tuning
process can stabilize the performance of large language models when
facing long text history (\eg precision and F1-score on
\texttt{Stack elec}). This shows the effectiveness of supervised
fine-tuning in enhancing the ability of LLMs to understand
sequential interaction contexts.

\end{document}